\documentclass[preprint]{elsarticle} 

\makeatletter
	\def\ps@pprintTitle{%
 	\let\@oddhead\@empty
	\let\@evenhead\@empty
	\def\@oddfoot{\centerline{\thepage}}%
	\let\@evenfoot\@oddfoot}
\makeatother

\usepackage{etoolbox}
\patchcmd{\MaketitleBox}{\footnotesize\itshape\elsaddress\par\vskip36pt}{\footnotesize\itshape\elsaddress\par\parbox[b][36pt]{\linewidth}{\vfill\hfill\textnormal{\today}\hfill\null\vfill}}{}{}%
\patchcmd{\pprintMaketitle}{\footnotesize\itshape\elsaddress\par\vskip36pt}{\footnotesize\itshape\elsaddress\par\parbox[b][36pt]{\linewidth}{\vfill\hfill\textnormal{\today}\hfill\null\vfill}}{}{}%

\usepackage[
            colorlinks=true,%
            breaklinks=true,%
            linkcolor=blue,
            urlcolor=blue,%
            citecolor=blue,%
            pdftitle={Toward Reliable Railway-Bogie Response Prediction Using Multifidelity TDNN and Physics-Informed Residual Learning}, 
            pdfkeywords={railway bogie dynamics, multifidelity learning, time-delay neural network, physics-informed residual learning, simulation--test correlation},	
            pdfauthor={Gyeolhee Lee, Moosun Kim, Taewook Kwon, Jaehun Kim, Changsung Jeon, Dongjin Lee},		
            bookmarksopen=false,
            pdfpagemode=UseNone]{hyperref}
            
\usepackage[margin = 1.2in]{geometry}
\usepackage[T1]{fontenc}
\usepackage[english]{babel}
\usepackage[latin1]{inputenc} 
\usepackage{mathtools}
\usepackage{amsmath}
\usepackage{amsfonts}
\usepackage{amsthm}
\usepackage{amssymb}
\usepackage{array}
\usepackage{longtable}
\usepackage{algorithm} 
\usepackage{algorithmicx}
\usepackage{algpseudocode}
\usepackage{subcaption}
\usepackage{siunitx}

\usepackage{color}
\usepackage{url}
\usepackage{cleveref}
\usepackage{graphicx}
\usepackage{breakcites}
\usepackage{soul} 
\usepackage{enumitem}
\usepackage[title,titletoc,toc]{appendix}

\usepackage{tikz}
\usetikzlibrary{arrows.meta,positioning,shapes.geometric}
\usepackage[textsize=tiny]{todonotes}
 
\usepackage{amsmath}
\usepackage{graphicx}
\usepackage{tikz}
\usetikzlibrary{positioning,arrows.meta,fit,backgrounds}

\newtheorem{problem}{Problem}

\graphicspath{{./figures/}}

\newcommand{\plannedplot}[2][0.13\textheight]{%
	\IfFileExists{figures/#2}{%
		\includegraphics[width=\linewidth]{#2}%
	}{%
		\begingroup
		\setlength{\fboxsep}{6pt}%
		\fbox{\parbox[c][#1][c]{0.90\linewidth}{\centering\small
			\textbf{Planned data plot}\\[3pt]
			\texttt{\detokenize{#2}}}}
		\endgroup
	}%
}

\begin{document}
	
	\begin{frontmatter}
		
			\title{Toward Reliable Railway-Bogie Response Prediction Using Multifidelity TDNN and Physics-Informed Residual Learning}
		
		
		\author[hyu]{Gyeolhee Lee\fnref{equal}}

		\author[krri]{Moosun Kim\corref{cor1}}
		\ead{mskim@krri.re.kr}

		\author[hyu_fm]{Taewook Kwon\fnref{equal}}

		\author[hyu]{Jaehun Kim}

        \author[krri]{Changsung Jeon}

		\author[hyu]{Dongjin Lee}
		
		\cortext[cor1]{Corresponding author}
		\fntext[equal]{Gyeolhee Lee and Taewook Kwon contributed equally to this work as co-first authors.}
		
		\address[hyu]{Department of Automotive Engineering, Hanyang University, 222 Wangsimni-ro, Seongdong-gu, Seoul 04763, Republic of Korea}

		\address[krri]{Korea Railroad Research Institute, 176 Cheoldobangmulgwan-ro, Uiwang-si, Gyeonggi-do 16105, Republic of Korea}

        \address[hyu_fm] {Department of Future Mobility, Hanyang University, 222 Wangsimni-ro, Seongdong-gu, Seoul 04763, Republic of Korea}
		
		\begin{abstract}
					Railway engineers need simulation models that predict vehicle responses across operating scenarios that cannot be tested exhaustively. Agreement with representative measurements provides essential evidence, but calibration at a limited set of conditions does not guarantee accuracy elsewhere. We present a multifidelity railway-bogie response-correction method that treats multibody simulation histories as low-fidelity information and roller-rig measurements as high-fidelity evidence. This method combines an experiment-anchored fidelity assignment with physics-informed discrepancy learning for multichannel bogie-response histories. A time-delay neural network (TDNN) represents the condition-dependent simulation trend, and development-fitted amplitude alignment defines the low-fidelity baseline. A residual-correction network then models the reproducible response component not explained by this baseline and adds it to the baseline. An effective dynamic-balance equation constrains the learned discrepancy by representing differences in inertia, damping, stiffness, and external forcing between the simulated and physical systems. The training objective combines this constraint with residual matching, temporal smoothness, and a combined channel-2 acceleration loss selected using displacement--acceleration consistency evidence. For the evaluated reconstruction case, the corrected response gives a mean coefficient of determination of 0.8197, a mean normalized root-mean-square error (NRMSE) of \SI{4.6055}\%, and a mean normalized mean absolute error (NMAE) of \SI{1.9297}\%. These results provide initial evidence of accurate response prediction at the held-out 385~km/h condition.
		\end{abstract}	
		
		\begin{keyword}
				Railway bogie dynamics \sep model discrepancy \sep simulation--test correlation \sep multifidelity learning \sep physics-informed residual learning \sep roller-rig testing
		\end{keyword}
		
\end{frontmatter}

\section*{Nomenclature}
\addcontentsline{toc}{section}{Nomenclature}
\begingroup
\scriptsize
\setlength{\tabcolsep}{4pt}
\renewcommand{\arraystretch}{1.00}
\begin{longtable}{@{}>{\raggedright\arraybackslash}p{0.18\textwidth}>{\raggedright\arraybackslash}p{0.58\textwidth}>{\raggedright\arraybackslash}p{0.14\textwidth}@{}}
	\hline
	Symbol & Definition & Unit/domain \\
	\hline
	\endfirsthead
	\multicolumn{3}{@{}r@{}}{\textit{Nomenclature (continued)}}\\
	\hline
	Symbol & Definition & Unit/domain \\
	\hline
	\endhead
	$\mathbb{R}$ & Set of real numbers & -- \\
	$\mathbb{R}^{+}$ & Set of strictly positive real numbers, $\mathbb{R}^{+}:=\{x\in\mathbb{R}\mid x>0\}$ & -- \\
    $\mathbb{Z}$ & Set of integers & -- \\
	$\mathbb{N}$ & Set of natural numbers including zero, $\mathbb{N}:=\{0,1,2,\ldots\}$ & -- \\
	$\mathrm{FW}$, $\mathrm{FB}$, $\mathrm{RW}$, $\mathrm{RB}$ & Front axlebox, front bogie frame, rear axlebox, and rear bogie frame & -- \\
	ML, TDNN & Machine learning and time-delay neural network & -- \\
	$Y$ & Signed lateral wheel--rail contact force at an evaluated wheel, according to the model force convention & kN \\
	$Q$ & Corresponding compressive vertical wheel load at the same wheel & kN \\
	$Y/Q$ & Dimensionless derailment quotient evaluated for $Q>0$ & -- \\
	$F_{\mathrm{ext}}$, $F_{0}$ & Force transmitted through the external rod-bush markers and internal preload force & N \\
	$k_{\mathrm{ser}}$, $k_{\mathrm{par}}$, $d_b$ & Rod-bush series stiffness, parallel stiffness, and relative deformation & N/m, N/m, m \\
	$\mathbf{q}^{\mathrm{s}}$, $\mathbf{q}^{\mathrm{e}}$ & Simulation and experimental displacement-response vectors & mm \\
	$\mathbf{x}$ & Operating-condition vector comprising speed, excitation amplitude, and excitation frequency & \si[per-mode=symbol]{\kilo\metre\per\hour}, deg, Hz \\
    $\boldsymbol{\xi}$ & PINN input vector comprising speed and harmonic phase features & -- \\
	$\mathbf{u}^{W}$, $\widehat{\mathbf{q}}^{\mathrm{ML}}$ & Input-history window and unaligned ML representation of the simulation response & mm \\
    $N_T$ & Total number of complete TDNN history windows in the development conditions & -- \\
    $s_{q,j}$ & Development-derived displacement scale for channel $j$ & mm \\
    $[\mathbf{I}_J]$ & $J\times J$ identity matrix & -- \\
	$\mathcal{F}_{\boldsymbol{\phi}}$, $\boldsymbol{\phi}$ & Temporal ML map and its trainable parameter vector & -- \\
	$L$, $n_{\ell}$ & Number of affine TDNN layers and width of layer $\ell$ & -- \\
	$[\mathbf{W}_{\ell}]$, $\mathbf{b}_{\ell}$, $\boldsymbol{\sigma}_{\ell}$ & Weight matrix, bias vector, and componentwise activation of layer $\ell$ & -- \\
	$\mathbf{a}^{(\ell)}$, $\mathbf{z}^{(\ell)}$ & Preactivation and output vectors of layer $\ell$ & -- \\
	$n_{\theta}$ & Number of trainable residual-network parameters & -- \\
	$\Delta t$, $W$, $J$, $p$ & Sampling interval, input-history length, number of measured channels, and number of reduced coordinates & s, --, --, -- \\
	$N_c$ & Number of paired samples retained for operating condition $c$ & -- \\
    $\rho$ & Amplitude factor & -- \\
	$\widetilde{\mathbf{q}}^{\mathrm{ML}}$ & Amplitude-aligned ML baseline response & mm \\
	$\boldsymbol{\delta}$, $\boldsymbol{\varepsilon}$ & Latent model discrepancy and measurement error & mm \\
	$\mathbf{r}^{\mathrm{e}}$ & Observed experimental--baseline residual & mm \\
	$\mathcal{G}_{\boldsymbol{\theta}}$, $\boldsymbol{\theta}$ & Reduced-coordinate correction map and trainable parameter vector & -- \\
	$\widehat{\boldsymbol{\delta}}^{y}_{\boldsymbol{\theta}}$, $\boldsymbol{\delta}^{y}$ & Learned and exact reduced-coordinate discrepancies & reduced-coordinate unit \\
	$\boldsymbol{\Delta}_{\boldsymbol{\theta}}$ & Learned sensor-space correction & mm \\
    $\boldsymbol{\theta}_\mathrm{PINN}$
    & Trainable parameter vector of the experiment-only PINN
    & -- \\
	$\widehat{\mathbf{q}}$ & Discrepancy-corrected response & mm \\
	$\mathbf{y}_{s}$, $\mathbf{y}_{r}$ & Nominal and noise-free physical reduced-coordinate vectors & coordinate dependent \\
	$[\mathbf{H}_{q}]$ & Linear reduced-coordinate-to-sensor observation matrix & sensor unit per reduced-coordinate unit \\
	$[\mathbf{M}_{s}]$, $[\mathbf{C}_{s}]$, $[\mathbf{K}_{s}]$ & Nominal reduced mass, damping, and stiffness matrices & coordinate dependent \\
	$[\mathbf{M}_{r}]$, $[\mathbf{C}_{r}]$, $[\mathbf{K}_{r}]$ & Noise-free physical reduced mass, damping, and stiffness matrices & coordinate dependent \\
	$[\Delta\mathbf{M}]$, $[\Delta\mathbf{C}]$, $[\Delta\mathbf{K}]$ & Effective reduced parameter-discrepancy matrices & coordinate dependent \\
	$\mathbf{f}_{s}$, $\mathbf{f}_{r}$, $\Delta\mathbf{f}$ & Nominal, physical, and discrepancy generalized-force vectors & coordinate-conjugate generalized-force unit \\
	$\mathbf{g}_{\boldsymbol{\theta}}$ & Effective dynamic residual & coordinate-conjugate generalized-force unit \\
	$D_{\Delta t}$, $D_{\Delta t}^{2}$ & Centered discrete first and second derivatives & \si{\per\second}, \si{\per\second\squared} \\
	$[\mathbf{S}_{q}]$, $[\mathbf{S}_{\kappa}]$, $[\mathbf{S}_{g}]$ & Positive diagonal scale matrices for displacement, correction curvature, and dynamic residual & mm, \si{\milli\metre\per\second\squared}, coordinate-conjugate generalized-force unit \\
	$s_{a,2}$ & Channel-2 acceleration scale & \si{\metre\per\second\squared} \\
	$\gamma_q$ & Conversion factor from millimetres to metres & \si{\metre\per\milli\metre} \\
    $\mathcal{L}_{\mathrm{res}},\,\mathcal{L}_{\mathrm{smooth}},     \,\mathcal{L}_{\mathrm{acc}}$ & Residual-matching, correction-smoothness, and combined channel-2 acceleration & -- \\
    $\mathcal{L}_{\mathrm{phys}}$ & Dynamic-residual losses & -- \\
	$\lambda_{\mathrm{res}}$, $\lambda_{\mathrm{smooth}}$, $\lambda_{a,2}$ & Nonnegative loss weights & -- \\
    $\lambda_{a,2}^{\mathrm{kin}},\,
    \lambda_{a,2}^{\mathrm{cons}},\,\lambda_{\mathrm{phys}}$
    & Nonnegative weights of the proposed correction loss terms
    & -- \\
	$R^2$, RMSE, MAE & Coefficient of determination, root-mean-square error, and mean absolute error & --, mm, mm \\
    $\mathrm{NRMSE},\,\mathrm{NMAE}$ & Normalized root-mean-square error, normalized mean absolute error & \%, \% \\
    $\widehat a_{c,2},\,\widehat a^{\mathrm h}_{c,2},\,a^{\mathrm e}_{c,2}$ & Displacement-derived, acceleration-head, and measured channel-2 accelerations & $\mathrm{m/s^2}$ \\
    $\mathcal{B}_c$ & Condition-dependent coherent-band projection operator for the constituent terms of $\mathcal{L}_{\mathrm{acc}}$ & -- \\
    $i^-_{\nu,c},\,i^+_{\nu,c},\,N_{\nu}$ & First and last valid sample indices and total sample count for loss label $\nu$ & -- \\
    $\widehat q^{\mathrm{PINN}}_{c,j}$ & Experiment-only PINN prediction for response channel $j$ & mm \\
    $g^{\mathrm{PINN}}_{c,j}$ & Effective channelwise SDOF residual of the experiment-only PINN & $\mathrm{m/s^2}$ \\
    $\omega_c,\,\omega_{n,j},\,\zeta_j$ & Excitation angular frequency, effective natural angular frequency, and damping ratio for the experiment-only PINN & $\mathrm{rad/s}$, $\mathrm{rad/s}$, -- \\
    $A_j(v_c),\,B_j(v_c)$ & Speed-conditioned harmonic forcing coefficients of the effective SDOF model & $\mathrm{m/s^2}$ \\
    $\mathcal{L}_{\mathrm{PINN}},\,\mathcal{L}_{\mathrm{data}},\,    \mathcal{L}_{\mathrm{smooth}}^{\mathrm{PINN}}$ & Total, displacement-data, displacement-smoothness losses of the experiment-only PINN & -- \\
    $\mathcal{L}_{\mathrm{SDOF}}$ & Effective-SDOF loss of the experiment-only PINN & -- \\
    $\lambda_{\mathrm{smooth}}^{\mathrm{PINN}},\,\lambda_{\mathrm{SDOF}}$ & Smoothness and effective-SDOF loss weights of the experiment-only PINN & -- \\
    $\widehat{\mathbf r}^{\mathrm{NN}}_c$ & Four-channel residual predicted by the TDNN-assisted residual network & mm \\
    $\boldsymbol{\eta}$ & Trainable parameter vector of the TDNN-assisted residual network & -- \\
    $\mathcal{L}_{\mathrm{resNN}}$ & Residual-matching loss of the TDNN-assisted residual network & -- \\
    $\mathcal{K},\,k$ & Prespecified finite set of candidate integer sample lags and a candidate lag & --, samples \\
    $\mathcal{I}_5(k),\,N_5(k)$ & Valid evaluation-index set for lag $k$ and its cardinality & -- \\
    $k_{\mathrm{oracle}}^{*}$ & Target-informed global lag selected by post-hoc oracle temporal alignment & samples \\
    $\widehat{\mathbf q}^{\mathrm{oracle}}_5$ & Oracle-aligned corrected response at the fifth operating condition & mm \\
    $\overline q^{\mathrm e}_{5,j},\,\Delta q^{\mathrm e}_{5,j}$ & Measured sample mean and measured response range of channel $j$ at the fifth condition & mm \\
	\hline
\end{longtable}
\endgroup

\section{Introduction} \label{sec:intro}

Railway engineers use vehicle-dynamics simulations to assess stability, ride, component loads, and safety-related responses across many operating and design conditions. Physical tests cannot cover all of these conditions. A reliable simulation can therefore screen scenarios, guide test selection, and reduce repeated development tests. It cannot replace the physical procedures required for safety, certification, or vehicle acceptance. Agreement with measurements under representative conditions is an important initial requirement, but it does not guarantee accuracy under untested conditions. Suspension and contact uncertainty, differences between numerical and roller-rig boundaries, sensor behavior, and signal processing all cause simulation--test discrepancies. The challenge is to use simulation for broad coverage and measurements for direct evidence.

Railway multibody-dynamics (MBD) models predict vehicle responses by representing the suspension, wheel--rail contact, and track or test-rig boundary. A locomotive model-acceptance procedure combined numerical checks, standards-based dynamic tests, and field evidence \cite{spiryagin2022locomotive}. A real-time scaled-bogie-rig study likewise retained experimental comparison when assessing a simplified MBD model of the physical rig \cite{shrestha2020realtime}. Scaled roller-rig tests reported in the Journal of Mechanical Science and Technology compared measured and simulated critical speeds across worn wheel profiles \cite{hur2009critical}. Time-domain parameter identification can improve agreement with calibration measurements \cite{kraft2013calibration}, and scaled-vehicle validation can reveal response modes omitted from the model \cite{urda2020scaled}. These results support response-specific validation over multiple conditions; calibration on one dataset does not establish predictive accuracy elsewhere \cite{polach2014criteria,lance2022verification}.

Machine-learning (ML) surrogates can accelerate railway-dynamics prediction, but their reliability depends on whether the training data cover the intended operating domain. One railway study trained a convolutional diagnostic network with MBD-generated bogie-frame accelerations and then evaluated it with track data, while identifying model accuracy and computational efficiency as continuing limitations \cite{ye2022suspension}. Another learned long-horizon, low-frequency vehicle--track responses from short MBD histories, but reported inaccurate high-frequency impact predictions and insufficient operating conditions and training samples \cite{ye2021mbsnet}. Hunting-instability prediction likewise required a spectral sample library and combined field measurements, rig tests, and simulations because unsafe states could not be collected in line tests and rig speeds were restricted \cite{chen2023hunting}. A recent railway physics-informed neural network (PINN) study identifies limited sensing, uncertain operating conditions, model generalization, and computational reliability as practical barriers \cite{han2026satpinn}. These findings motivate a method that uses simulation to supply a condition-dependent nominal trend and the available measurements to learn the reproducible component of the observed simulation--test residual.

Multifidelity modeling provides this connection. Classical variable-fidelity methods use an inexpensive low-fidelity prediction to establish the trend of a surrogate for a more expensive high-fidelity response \cite{han2012hierarchical}. Neural approaches can instead learn the discrepancy between fidelity levels directly \cite{zhang2022physicsInformed,davis2025residual}. Railway applications use different fidelity definitions and prediction targets. Prior work combined a simplified moving-load model with a detailed bridge--train interaction model for railway-bridge reliability \cite{hirzinger2024railway}, and another study combined reduced- and full-order models to predict railway-bridge sensor responses \cite{torzoni2023railwayBridge}. At the vehicle level, recent work fused MBD and finite-element simulations for high-speed-train crash-response prediction \cite{xie2025crash}. These studies combine numerical models of different fidelity; they do not address experiment-informed correction of multichannel bogie-response histories.

In this work, we present a simulation-informed method for correcting four bogie displacement histories measured in high-speed roller-rig tests under harmonic yaw excitation. The central methodological contribution is the combination of experiment-anchored multifidelity learning and physics-informed model-discrepancy correction for multichannel bogie-response histories. Within this combined formulation, MBD histories provide the low-fidelity response trend, physical measurements provide high-fidelity evidence of the remaining discrepancy, and an effective mass--damping--stiffness balance constrains the learned correction. We realize this contribution through: (1) a TDNN baseline with development-fitted amplitude alignment; (2) a response decomposition that distinguishes the aligned baseline, latent model discrepancy, and measurement error; (3) a correction objective that combines residual matching, temporal smoothness, the effective dynamic residual, and a combined channel-2 acceleration loss selected using displacement--acceleration consistency evidence; and (4) an assessment structure that separates full-vehicle model-consistency evidence from roller-rig response-reconstruction evidence. 

Section~\ref{sec:bogie_configuration} presents the test configuration, measured responses, multibody model, and the TDNN and multifidelity foundations used in this study. Section~\ref{sec:method} defines the proposed response decomposition, TDNN-based ML baseline, alignment, model-discrepancy correction, and training and inference workflow. Section~\ref{sec:results} presents the multibody-model assessment, the learning-based comparisons, and the discrepancy-corrected response. The appendices document the full-vehicle inputs and displacement--acceleration channel-consistency study.

\section{Roller-Rig Test Configuration and Modeling Foundations} \label{sec:bogie_configuration}

We first define the tested conditions and then describe the test rig, numerical models, and modeling concepts used in the proposed correction method. The roller-rig records support comparisons only for these conditions; they do not define the full range in which the model may be used.

\subsection{Test Vehicle, Operating Conditions, and Roller-Rig Boundary} \label{sec:test_configuration}\label{sec:rig_boundary}

\begin{figure}[H]
	\centering
	\begin{subfigure}[t]{0.36\textwidth}
		\centering
		\includegraphics[width=\linewidth]{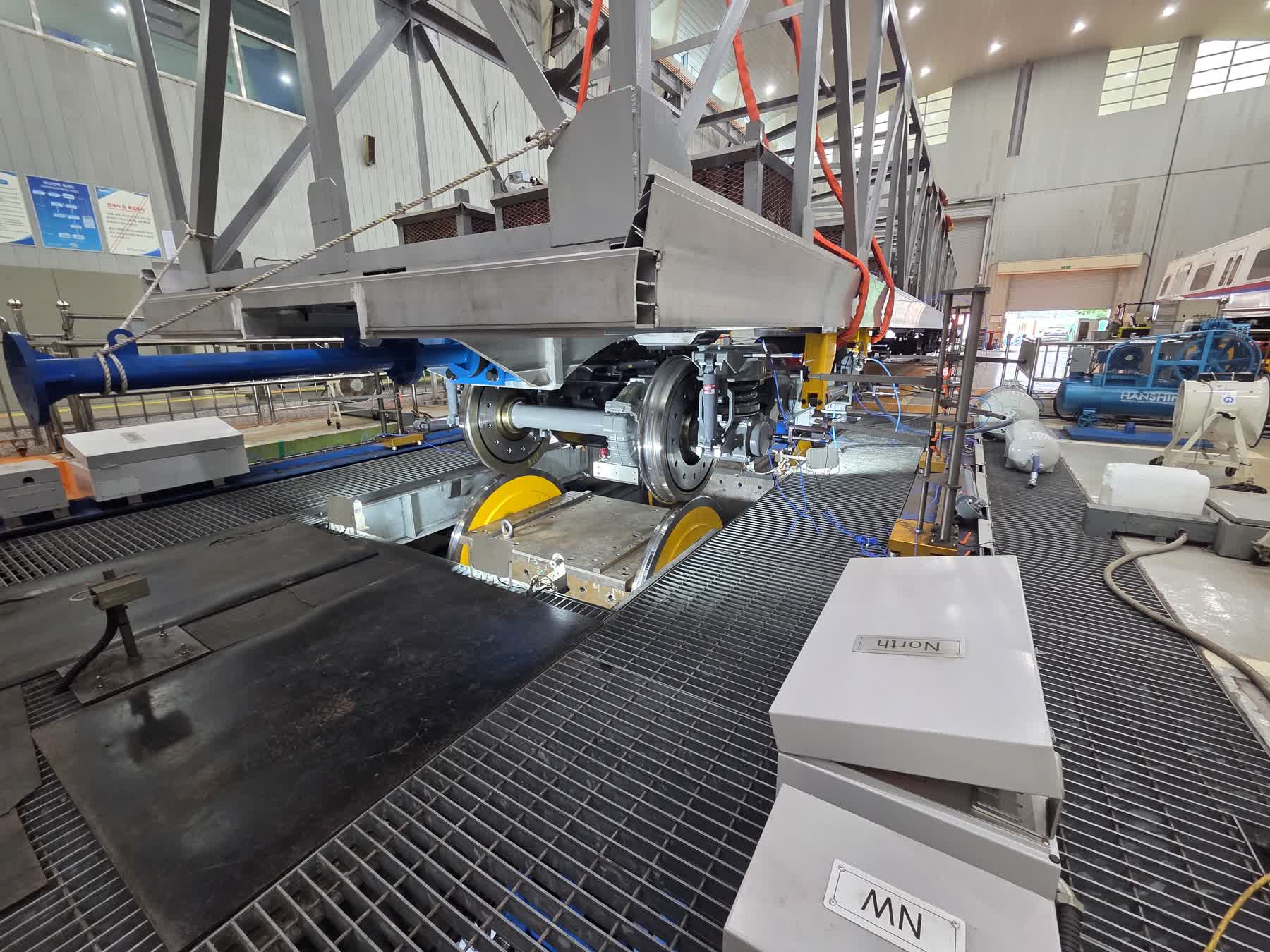}
		\caption{Test installation.}
	\end{subfigure}
	\hfill
	\begin{subfigure}[t]{0.62\textwidth}
		\centering
		\includegraphics[width=\linewidth]{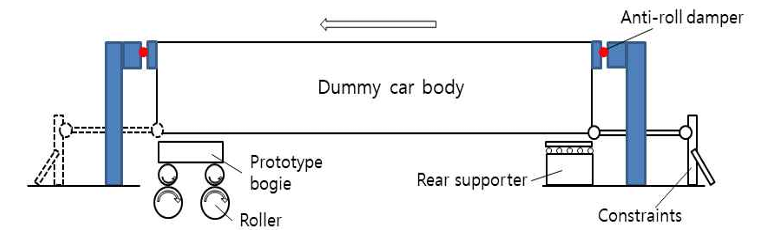}
		\caption{Mechanical boundary representation.}
	\end{subfigure}
	\caption{Semi-full-car roller-rig configuration: (a) prototype-bogie test installation; (b) modeled boundary, supports, constraints, and external anti-roll dampers.}
	\label{fig:rig_configuration}
\end{figure}

We distinguish the bogie suspension definition, the roller-rig model used for response correlation, and the full-vehicle model used for running-stability assessment. The roller-rig model represents the semi-full-car installation by coupling prototype bogie WJH-IHB No.~M001 to two roller pairs, a dummy car body, a rear support, a ball-joint restraint, and external anti-roll dampers. The physical rig is rated to \SI[per-mode=symbol]{420}{\kilo\metre\per\hour}, and the present tests extend to \SI[per-mode=symbol]{385}{\kilo\metre\per\hour}. The full-vehicle model instead contains one car body, two bogies, and four wheelsets. The configurations share bogie-level component definitions but retain distinct boundaries and evidentiary roles. Figure~\ref{fig:rig_configuration} illustrates the physical installation and its mechanical representation.

All configurations use the $x$-axis in the running direction, the $y$-axis laterally to the right, and the $z$-axis vertically upward. The car-body reference lies on the rail plane midway between the bogie centers, and each bogie reference lies on the rail plane at its center. Response correlation additionally requires matched quantities, units, axes, signs, reference frames, time bases, and excitation metadata.

We evaluate five nominal speed conditions: 300, 320, 340, 360, and \SI[per-mode=symbol]{385}{\kilo\metre\per\hour}. The intended split uses 300--\SI[per-mode=symbol]{360}{\kilo\metre\per\hour} for development and \SI[per-mode=symbol]{385}{\kilo\metre\per\hour} for evaluation. The final condition is independent only if every target-informed operation excludes it. Table~\ref{tab:test_configuration} summarizes the test conditions and data roles.

\begin{table}[htbp]
	\centering
	\caption{Roller-rig test conditions and data roles used in this study.}
	\label{tab:test_configuration}
	\begin{tabular}{>{\raggedright\arraybackslash}p{0.24\textwidth}p{0.67\textwidth}}
		\hline
		Item & Description \\
		\hline
		Test article and rig & Prototype bogie WJH-IHB No.~M001 in a semi-full-car installation; two roller sets; rig rating of \SI[per-mode=symbol]{420}{\kilo\metre\per\hour} and test maximum of \SI[per-mode=symbol]{385}{\kilo\metre\per\hour} \\
		Wheel--rail interface & PT04 wheel profile and 60E1 rail profile at a 1:20 inclination \\
		Roller-yaw excitation & Peak angular amplitude of \SI{0.03}{\degree} from the neutral position at \SI{2}{\hertz} \\
		Residual-correction speed set & 300, 320, 340, 360, and \SI[per-mode=symbol]{385}{\kilo\metre\per\hour} \\
		Measured channels and acquisition & Four lateral LVDT displacement channels and 16 lateral and vertical accelerometer channels; the unexcited runs were recorded at \SI{1}{\kilo\hertz} \\
		\hline
	\end{tabular}
\end{table}

In the roller-rig simulation, we prescribe a harmonic yaw rotation about the roller model's vertical axis for approximately 10~s, with a peak angular amplitude of \SI{0.03}{\degree} measured from the neutral position and a frequency of \SI{2}{\hertz}. The physical test retains at least 10~s of data before and after this input. For the rig geometry, this peak rotation produces peak horizontal displacements of approximately \SI{0.73}{\milli\metre} at the actuator attachment point and \SI{0.65}{\milli\metre} at the wheelset center. The actuator attachment point lies approximately \SI{1400}{\milli\metre} from the motion-platform rotation center. The wheelset center lies approximately \SI{1250}{\milli\metre} from the bogie center. A change-point procedure selects a steady forced-response subsegment at each speed, so the learning workflow does not assume that the complete commanded interval is suitable. Reproduction requires the selected steady-response interval for every condition.

\subsection{Instrumentation and Response Data} \label{sec:experiment}

\begin{figure}[H]
	\centering
	\begin{subfigure}[t]{0.42\textwidth}
		\centering
		\includegraphics[width=\linewidth]{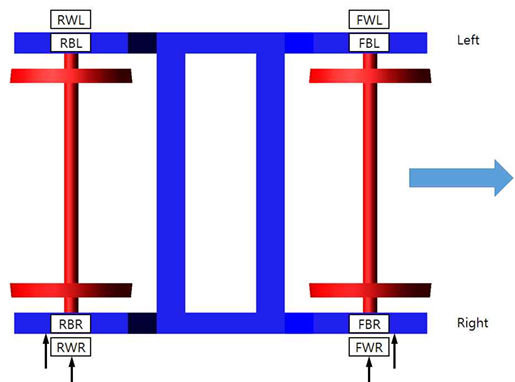}
		\caption{Measured sensor locations.}
	\end{subfigure}
	\hfill
	\begin{subfigure}[t]{0.55\textwidth}
		\centering
		\includegraphics[width=\linewidth]{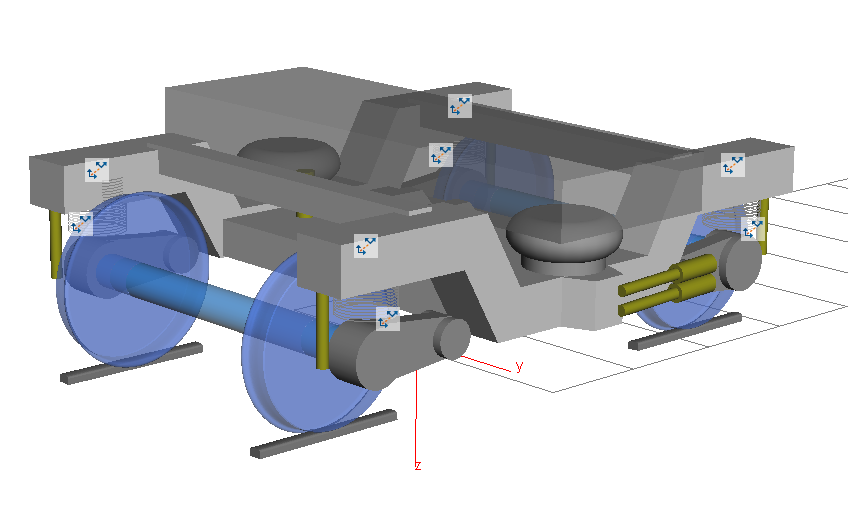}
		\caption{Virtual output locations.}
	\end{subfigure}
	\caption{Measured and virtual response locations. The response-correction targets are indexed by the FW, FB, RW, and RB locations; physical-axis and sign correspondence require confirmation before channelwise pairing.}
	\label{fig:sensor_correspondence}
\end{figure}

Figure~\ref{fig:sensor_correspondence} illustrates the measured sensor locations and the corresponding virtual outputs used to define the response channels.

The roller-rig dataset contains 20 measured channels: four linear-variable differential transformer (LVDT) displacement channels and 16 channels of lateral and vertical acceleration at front/rear axlebox and bogie-frame locations on both sides of the instrumented bogie. The LVDTs have a \SI{\pm 50}{\milli\metre} measurement range. The axlebox and bogie-frame accelerometers have nominal ranges of \SI{50}{g} and \SI{10}{g}. The test system recorded the unexcited runs at \SI{1}{\kilo\hertz}. The raw sampling rate and any resampling applied to each excited learning record must be confirmed from its acquisition and preprocessing metadata. We order any four-channel displacement response as
\begin{equation}
	\mathbf{q}(t)
	=
	\bigl(q_{\mathrm{FW}}(t),q_{\mathrm{FB}}(t),q_{\mathrm{RW}}(t),q_{\mathrm{RB}}(t)\bigr)^{\mathsf T}
	\equiv
	\bigl(q_1(t),q_2(t),q_3(t),q_4(t)\bigr)^{\mathsf T},
	\label{eq:response_vector}
\end{equation}
where $t\in\mathbb{R}^{+}$ is physical time, $\mathbf{q}(t)\in\mathbb{R}^{4}$ is a generic displacement-response vector, and $q_j(t)$ is its $j$th channel. The label $\mathrm{FW}$ denotes the front axlebox, $\mathrm{FB}$ denotes the front bogie frame, $\mathrm{RW}$ denotes the rear axlebox, and $\mathrm{RB}$ denotes the rear bogie frame. We use location-based channel symbols because the physical response axis has not been verified consistently for every measured-to-virtual pairing. Reproducible pairing also requires the positive direction, reference frame, unit, acquisition time base, and corresponding virtual output.

Each excited time history follows a common protocol with at least 10~s before the input, approximately 10~s of harmonic excitation, and at least 10~s after the input. We apply a moving-RMS change-point procedure to identify a steady forced-response subsegment within the commanded interval. We then place the corresponding simulation and experimental segments on a common time base before applying amplitude alignment and development-derived normalization.

\begin{figure}[H]
	\centering
	\includegraphics[width=0.64\linewidth]{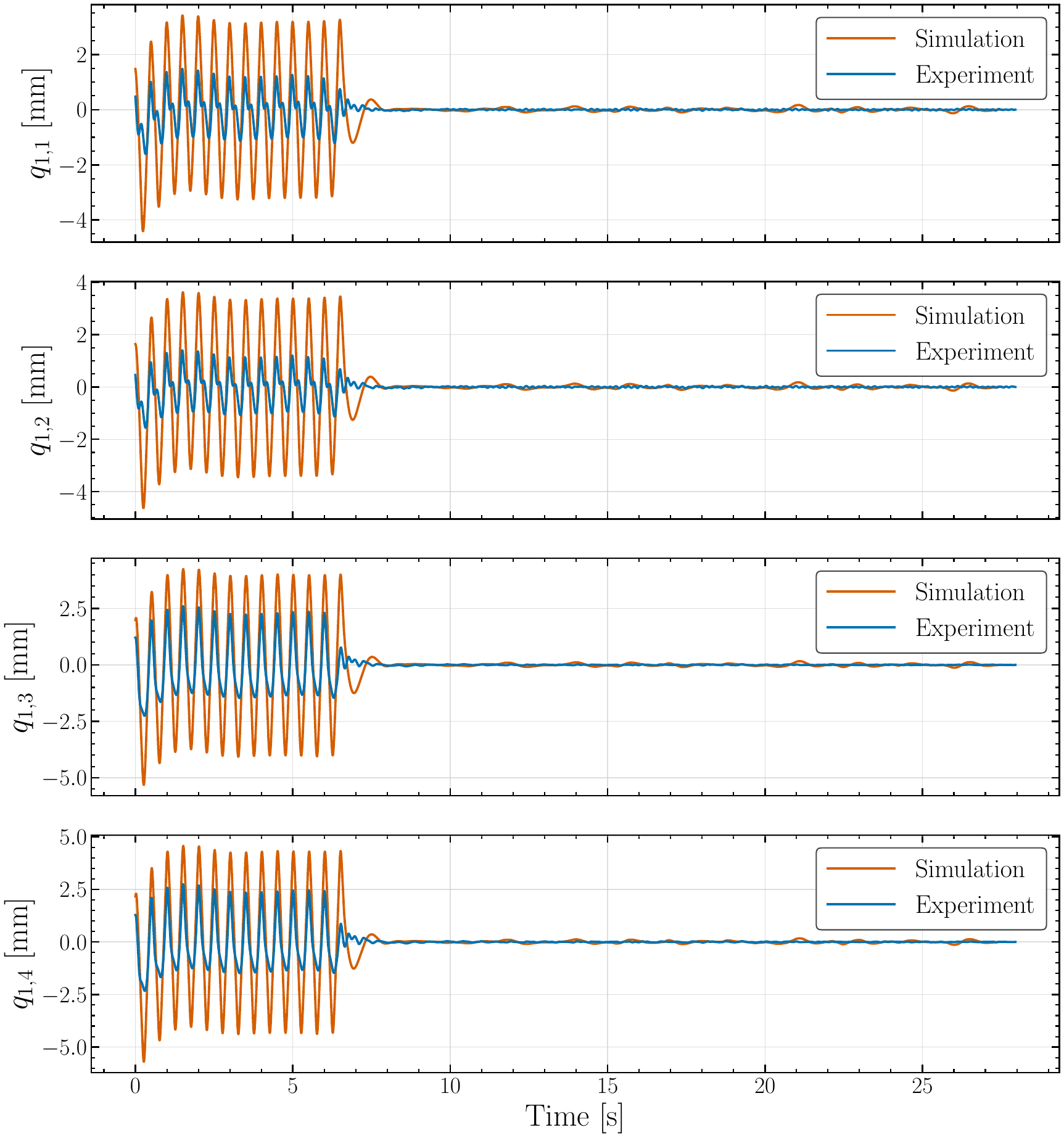}
	\caption{Representative experimental and simulation displacement responses at \SI[per-mode=symbol]{300}{\kilo\metre\per\hour} after interval extraction. The displayed records continue through the post-excitation response.}
	\label{fig:data_preparation}
\end{figure}

Let $c=1,\ldots,5$ index the five speed conditions in ascending order. The intended split uses $c=1,\ldots,4$ for development and $c=5$ for evaluation. For condition $c$, let $i=0,\ldots,N_c-1$ index the $N_c$ paired samples retained after fixed preprocessing, where $N_c\in\mathbb{N}$ and $N_c>0$. The common time grid is $t_i=t_0+i\Delta t$, with $t_0\in\mathbb{R}$ and $\Delta t\in\mathbb{R}^{+}$. Let $j=1,\ldots,J$ index the $J=4$ displacement channels. Thus, $c$, $i$, $j$, $N_c$, and $J$ are natural-number indices or counts. We collect the paired simulation and experimental responses as
\begin{equation}
\begin{aligned}
	\mathbf{q}^{\mathrm{s}}_c(t_i)
	&:=\bigl(q^{\mathrm{s}}_{c,1}(t_i),\ldots,q^{\mathrm{s}}_{c,J}(t_i)\bigr)^{\mathsf T}\in\mathbb{R}^{J},\\
	\mathbf{q}^{\mathrm{e}}_c(t_i)
	&:=\bigl(q^{\mathrm{e}}_{c,1}(t_i),\ldots,q^{\mathrm{e}}_{c,J}(t_i)\bigr)^{\mathsf T}\in\mathbb{R}^{J}.
\end{aligned}
\label{eq:response_source_definitions}
\end{equation}
Figure~\ref{fig:data_preparation} presents representative realizations of $\mathbf{q}^{\mathrm{s}}_c$ and $\mathbf{q}^{\mathrm{e}}_c$ at \SI[per-mode=symbol]{300}{\kilo\metre\per\hour} after interval extraction. The figure illustrates the paired signal format and the remaining response difference before discrepancy correction; it does not establish model validity or quantify agreement.

The operating-condition input is $\mathbf{x}_c:=\bigl(v_c,\alpha_c,f_c\bigr)^{\mathsf T}\in\mathbb{R}^{3}$. The upright superscript $\mathrm{s}$ labels simulation quantities, and $\mathrm{e}$ labels experimental quantities; they are not scalar variables. The response vectors use millimetres. The scalar $v_c\in\mathbb{R}$ is the roller speed (\si[per-mode=symbol]{\kilo\metre\per\hour}), $\alpha_c\in\mathbb{R}$ is the peak yaw-excitation amplitude measured from the neutral position (\si{\degree}), and $f_c\in\mathbb{R}$ is the excitation frequency (\si{\hertz}).

For the PINN-based models, we define the time-dependent input vector\[\boldsymbol{\xi}_c(t_i):=\bigl(v_c,\,\sin(2\pi f_c t_i),\,\cos(2\pi f_c t_i)\bigr)^{\mathsf T}\in\mathbb{R}^{3}.\] The phase features are used in the experiment-only PINN and the proposed physics-informed residual model, but not in the TDNN baseline. Thus, the current PINN implementation uses $v_c$ directly and $f_c$ through the harmonic phase features, while direct inclusion of $\alpha_c$ and $f_c$ in $\mathbf{x}_c$ remains a proposed extension for excitation-conditioned prediction.

The intended evaluation uses 300--\SI[per-mode=symbol]{360}{\kilo\metre\per\hour} for development and \SI[per-mode=symbol]{385}{\kilo\metre\per\hour} for evaluation. Until the sensor dictionary is resolved, the channelwise metrics compare provisionally paired response arrays rather than physically validated sensor-to-model responses.

\subsection{Physics-Based Vehicle Model and Simulation Implementation} \label{sec:simulation_model}

We built two related multibody configurations using SIMULIA Simpack 2025x.2 \cite{dassault2024simpack2025x}: a roller-rig model for simulation--test response correction and a full-vehicle model for consistency assessment. The roller-rig configuration represents the tested running gear with one bogie frame and two wheelsets. Two roller pairs support the two wheelsets and provide the rolling boundary. A rigid dummy car body carries the imposed vertical load. Its geometry and support reactions distribute this load between the pair of secondary-suspension elements above the test bogie and the rear support. The ball-joint restraint and external anti-roll dampers complete the semi-full-car boundary defined in Section~\ref{sec:rig_boundary}. This reduced representation preserves the rig load path without introducing the second bogie of the full vehicle. The full-vehicle model contains one car body, two bogies, and four wheelsets. We model these principal assemblies as rigid bodies whose mass centers and inertia tensors are assigned in their local reference frames.

\begin{figure}[H]
	\centering
	\begin{subfigure}[t]{0.44\textwidth}
		\centering
		\includegraphics[width=\linewidth]{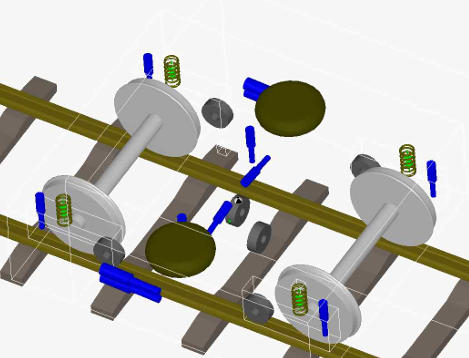}
		\caption{Suspension and force-element topology.}
	\end{subfigure}
	\hfill
	\begin{subfigure}[t]{0.5\textwidth}
		\centering
		\includegraphics[width=\linewidth]{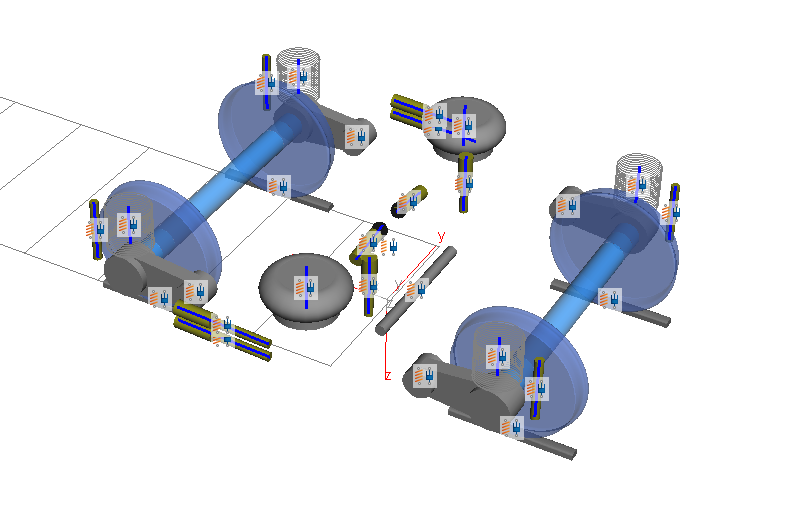
        }
		\caption{One-bogie, two-wheelset configuration.}
	\end{subfigure}
	\caption{Bogie-level model definition shared by the related multibody configurations: (a) implemented suspension and force-element topology; (b) assembled one-bogie, two-wheelset configuration used to define the component arrangement. The external roller-rig and full-vehicle boundaries are defined separately.}
	\label{fig:bogie_model}
\end{figure}

Figure~\ref{fig:bogie_model} illustrates the force-element topology and assembled bogie used to instantiate these connections.

The full-vehicle geometry uses a \SI{2.600}{\metre} bogie wheelbase and a \SI{17.3}{\metre} longitudinal distance between bogie centers. The lateral distance between the air-spring centers is \SI{2.100}{\metre}, the track gauge is \SI{1.435}{\metre}, and the wheel back-to-back distance is \SI{1.356}{\metre} with a tolerance of +0/\SI{-0.002}{\metre}. The new-wheel diameter is \SI{0.860}{\metre}, the worn-wheel diameter is \SI{0.760}{\metre}, and the nominal distance between wheel--rail contact points is \SI{1.500}{\metre}. \ref{app:model_parameters} provides the reported rigid-body properties and selected implementation details of the full-vehicle model. Some detailed suspension and connection parameters are not reported due to confidentiality and security restrictions.

The suspension model follows the physical load path from each wheelset to the bogie frame and then to the supported body. In the primary stage, a double-coil spring pair carries the axlebox-to-frame elastic load, an axlebox rod bush represents the compliant guidance connection, and a vertical oil damper dissipates relative vertical motion. We implement the primary vertical damper with a serial point-to-point spring--damper element; the series stiffness represents attachment compliance rather than an additional physical suspension stage. The rod-bush element retains its compliant internal-node structure, so the solver's internal preload and the force transmitted between its external markers are not generally identical.

The secondary stage transfers load between the bogie frame and the body through the air springs, vertical and lateral oil dampers, yaw dampers, anti-roll bar, and center pivot. The equivalent-air-spring model does not include pneumatic state dynamics. We therefore use equivalent force elements with prescribed translational stiffnesses in the longitudinal, lateral, and vertical directions for the empty-car condition and treat the omitted pneumatic dynamics as model-form uncertainty. We implement the secondary vertical, lateral, and yaw dampers as serial point-to-point spring--damper elements. The lateral and yaw elements use prescribed nonlinear velocity--force relations, while their series stiffnesses represent mounting compliance. A rotational spring element represents the anti-roll bar. The center pivot is represented by an equivalent force element that primarily provides longitudinal stiffness. The force element is connected through a rigidly attached dummy bolster to transfer the pivot load to the body. This construction transfers the pivot load without introducing an unintended flexible body mode. We replicate every suspension element at its specified left/right and leading/trailing attachment locations. Table~\ref{tab:component_implementation} presents the physical-to-numerical mapping, while~\ref{app:model_parameters} provides selected supporting model details.

\begin{table}[htbp]
	\centering
	\caption{Physical-to-numerical mapping of the shared bogie suspension model.}
	\label{tab:component_implementation}
	\small
	\begin{tabular}{@{}>{\raggedright\arraybackslash}p{0.18\textwidth}>{\raggedright\arraybackslash}p{0.31\textwidth}>{\raggedright\arraybackslash}p{0.41\textwidth}@{}}
		\hline
		Physical component & Mechanical connection & Multibody representation \\
		\hline
		Primary coil-spring pair & Wheelset/axlebox to bogie frame & Shear-spring element with vertical and two shear stiffness components \\
		Axlebox rod bush & Compliant axlebox-to-frame guidance & Bushing force element with internal compliance; external transmitted force distinguished from internal preload \\
		Primary vertical damper & Wheelset/axlebox to bogie frame & Serial point-to-point spring--damper element \\
		Secondary air spring & Bogie frame to supported body & Equivalent force element based on the specified static load--stiffness relation \\
		Secondary vertical, lateral, and yaw dampers & Bogie frame to supported body & Serial point-to-point spring--damper elements; nonlinear velocity--force laws retained for the lateral and yaw dampers \\
		Anti-roll bar & Bogie-frame/body roll connection & Rotational spring element \\
		Center pivot & Bogie frame to body center & Force element connected through a rigidly attached dummy bolster \\
		\hline
	\end{tabular}
\end{table}

\begin{figure}[htbp]
\centering
\includegraphics[width=0.8\linewidth]{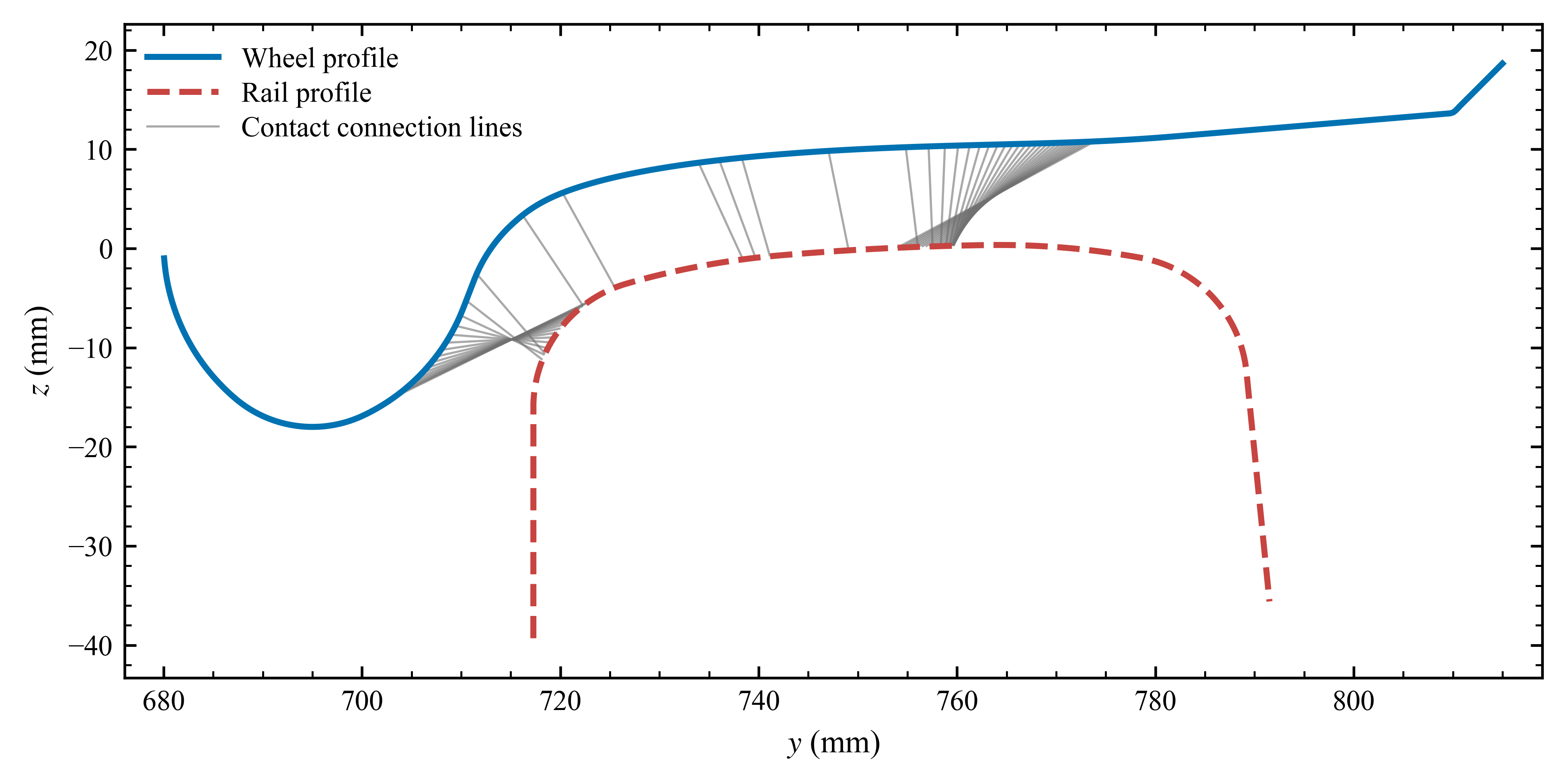}
\caption{Representative wheel--rail contact geometry used in the multibody models. The PT04 wheel and UIC60-20 rail profiles are shown together with the contact locations identified by the contact model.}
\label{fig:wheel_rail_contact}
\end{figure}

The multibody models use the PT04 wheel profile and the UIC60-20 rail profile with a profile-based contact formulation. Fig.~\ref{fig:wheel_rail_contact} shows a representative wheel--rail contact geometry obtained from the implemented contact model.

For the preload assessment, we compare the total transmitted spring forces, including the initial-deformation contributions, and use externally transmitted marker forces for elements with internal nodes.

For the roller-rig simulation, we prescribe the roller-speed and yaw-excitation cases, apply the support constraints, and place virtual displacement and acceleration outputs on the corresponding axlebox and bogie-frame components. We pair each virtual output with a measured channel only after matching the quantity of interest, unit, axis, positive direction, reference frame, time base, excitation schedule, and operating-condition metadata. Although we use Simpack in this study, the correction method accepts compatible time histories from another railway multibody dynamics solver.

\begin{problem}
Multibody simulations use idealized components, connections, and boundary conditions, which can produce a model discrepancy between simulated and experimental responses. We seek to correct this discrepancy while preserving the physical trends provided by the simulation.
\label{prob:model_discrepancy}
\end{problem}

\subsection{Simulation-Model Consistency Assessment} \label{sec:model_consistency}

We examine the separate full-vehicle configuration through three consistency assessments that use bogie-level component definitions related to those in the rig model. First, we assess the preload solution through force-element convergence, residual constraint acceleration, and comparison of principal spring, elastic-joint, and axle-load quantities with an independent static-equilibrium calculation. Second, a nonlinear deceleration scenario tracks the lateral oscillation amplitudes of the car body and bogie frames to exercise the hunting-stability transition and critical-speed procedure. Third, we process selected wheel--rail lateral force, wheel-unloading ratio, and derailment quotient ($Y/Q$) using EN~14363-based filtering, spatial averaging, and upper-tail extraction \cite{en14363_2022}, and compare the results descriptively with available cross-code outputs. At each evaluated wheel and time $t$, $Y(t)\in\mathbb{R}$ denotes the signed lateral wheel--rail contact force in kN under the model force convention. The quantity $Q(t)\in\mathbb{R}^{+}$ denotes the corresponding compressive vertical wheel load in kN. The derailment quotient $Y(t)/Q(t)\in\mathbb{R}$ is dimensionless. These assessments do not establish EN~14363 compliance or vehicle acceptance.

Together, these assessments characterize the assembly and numerical behavior of the full-vehicle implementation and provide limited cross-code context. Section~\ref{sec:simulation_results} presents the corresponding findings and limitations. The next two subsections introduce the temporal representation and multifidelity concepts used to connect these simulation histories with the roller-rig measurements.

\begin{figure}[t!]
	\centering
	\begin{tikzpicture}[
		>=Latex,
		font=\small,
		data/.style={draw, rounded corners=2pt, align=center, inner sep=4pt, font=\small, fill=white},
		block/.style={draw, rounded corners=2pt, align=center, inner sep=4pt, font=\small, fill=black!4},
		desc/.style={align=center, font=\footnotesize, inner sep=2pt},
		neuron/.style={circle, draw=black!75, line width=0.6pt, minimum size=5.2mm, inner sep=0pt, fill=white},
		neuron_in/.style={neuron, fill=blue!10, draw=blue!70!black},
		neuron_hid/.style={neuron, fill=teal!10, draw=teal!70!black},
		neuron_out/.style={neuron, fill=orange!12, draw=orange!80!black},
		flow/.style={->, line width=0.65pt},
		fitflow/.style={->, dashed, line width=0.65pt},
		conn/.style={-, draw=black!25, line width=0.35pt}
	]
		\node[data, text width=2.8cm] (history) at (-3.2, 0.7) {
			tapped simulation\\response\\
			$\mathbf{q}^{\mathrm{s}}_c(t_i)$\\
			$\mathbf{q}^{\mathrm{s}}_c(t_{i-1})$\\
			$\vdots$\\
			$\mathbf{q}^{\mathrm{s}}_c(t_{i-W+1})$
		};
		\node[data, text width=2.8cm] (condition) at (-3.2, -1.2) {
			speed condition\\
			$v_c$
		};

		\node[neuron_in] (in-1) at (0.0, 1.0) {};
		\node[neuron_in] (in-2) at (0.0, 0.4) {};
		\node (in-dots) at (0.0, -0.06) {$\vdots$};
		\node[neuron_in] (in-3) at (0.0, -0.7) {};
		\node[neuron_in] (in-4) at (0.0, -1.3) {};

		\node[neuron_hid] (h1-1) at (2.4, 1.2) {};
		\node[neuron_hid] (h1-2) at (2.4, 0.5) {};
		\node (h1-dots) at (2.4, -0.06) {$\vdots$};
		\node[neuron_hid] (h1-3) at (2.4, -0.8) {};
		\node[neuron_hid] (h1-4) at (2.4, -1.5) {};

		\node at (3.7, 1.2) {$\cdots$};
		\node at (3.7, 0.5) {$\cdots$};
		\node at (3.7, -0.8) {$\cdots$};
		\node at (3.7, -1.5) {$\cdots$};

		\node[neuron_hid] (h2-1) at (5.0, 1.2) {};
		\node[neuron_hid] (h2-2) at (5.0, 0.5) {};
		\node (h2-dots) at (5.0, -0.06) {$\vdots$};
		\node[neuron_hid] (h2-3) at (5.0, -0.8) {};
		\node[neuron_hid] (h2-4) at (5.0, -1.5) {};

		\node[neuron_out] (out-1) at (7.6, 0.6) {};
		\node (out-dots) at (7.6, -0.1) {$\vdots$};
		\node[neuron_out] (out-2) at (7.6, -0.9) {};

		\foreach \i in {1,2,3,4} {
			\foreach \j in {1,2,3,4} {
				\draw[conn] (in-\i) -- (h1-\j);
			}
		}
		\foreach \i in {1,2,3,4} {
			\foreach \j in {1,2} {
				\draw[conn] (h2-\i) -- (out-\j);
			}
		}

		\node[desc, anchor=north, text width=2.5cm] (desc-in) at (0.0, -2.0) {
			\textbf{input layer}\\
			$\mathbf{z}^{(0)}_{c,i}\in\mathbb{R}^{JW+1}$
		};
		\node[desc, anchor=north, text width=4.6cm] (desc-hid) at (3.7, -2.0) {
			\textbf{hidden layers} $\ell=1,\ldots,L-1$\\
			$\mathbf{a}^{(\ell)}_{c,i}=\mathbf{W}_{\ell}\mathbf{z}^{(\ell-1)}_{c,i}+\mathbf{b}_{\ell}$\\[2pt]
			$\mathbf{z}^{(\ell)}_{c,i}=\boldsymbol{\sigma}_{\ell}(\mathbf{a}^{(\ell)}_{c,i})$
		};
		\node[desc, anchor=north, text width=3.8cm] (desc-out) at (7.6, -2.0) {
			\textbf{output layer} $L$\\
			$\mathbf{a}^{(L)}_{c,i}=\mathbf{W}_L\mathbf{z}^{(L-1)}_{c,i}+\mathbf{b}_L$\\[2pt]
			$\widehat{\mathbf{q}}^{\mathrm{ML}}_c(t_i;\boldsymbol{\phi})=\boldsymbol{\sigma}_L(\mathbf{a}^{(L)}_{c,i})$
		};

		\node[block, text width=2.6cm] (loss) at (7.6, 2.3) {
			development loss\\
			Section~\ref{sec:tdnn}
		};
		\node[data, text width=2.6cm] (target) at (7.6, 3.7) {
			simulation target\\
			$\mathbf{q}^{\mathrm{s}}_c(t_i)$
		};

		\draw[flow] (history.east) -- node[above, font=\scriptsize] {$\mathbf{u}_{c,i}^{W}$} (in-1.west);
		\draw[flow] (history.east) -- (in-2.west);
		\draw[flow] (condition.east) -- (in-3.west);
		\draw[flow] (condition.east) -- (in-4.west);

		\draw[flow] (out-1.north) -- (loss.south);

		\draw[flow] (target) -- (loss);

		\draw[fitflow] (loss.west) to[out=180,in=90] 
			node[above, font=\scriptsize] {update $\boldsymbol{\phi}$} (3.7, 1.8);

	\end{tikzpicture}
	\caption{Detailed finite-history TDNN used to construct the ML representation of the simulation response. The $W$ tapped response vectors and speed condition $v_c$ form the augmented input $\mathbf{z}^{(0)}_{c,i}$. Each layer applies an affine transformation followed by a componentwise activation; the last layer returns the unaligned ML response. Solid arrows show forward and loss-evaluation paths, and the dashed arrow denotes development-only parameter fitting. Experimental responses do not enter the TDNN stage.}
	\label{fig:tdnn_architecture}
\end{figure}
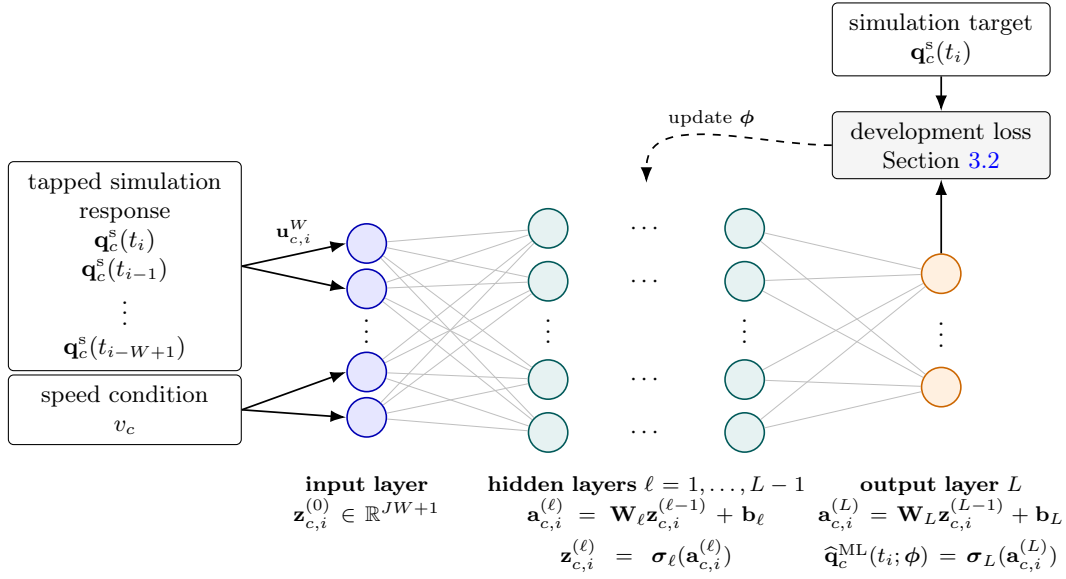

\subsection{Time-Delay Neural Network} \label{sec:tdnn_background}

A time-delay neural network (TDNN) represents temporal context explicitly through an input window of delayed samples, allowing its output to depend on a finite local history \cite{waibel1989phoneme}. The window length sets the represented memory: a short window can omit relevant response history, whereas a longer window increases the input dimension and data requirement.

For each operating condition $c=1,\ldots,5$, let $N_c\in\mathbb{N}$, with $N_c>0$, denote the number of paired samples retained after preprocessing. Let $W\in\mathbb{N}$ denote the TDNN history-window length in samples. Thus, each TDNN input contains the current $J$-channel simulation response and the preceding $W-1$ response samples. We require $1\leq W\leq\min_{c=1,\ldots,5}N_c$ so that a complete history window exists in every retained condition. Let $L\in\mathbb{N}$, with $L\geq2$, denote the number of affine layers, including the output layer. The positive integers $n_1,\ldots,n_{L-1}$ denote the hidden-layer widths. For compact layerwise notation, $n_0:=JW+1$ is the input width, and $n_L:=J$ is the output width. For $i=W-1,\ldots,N_c-1$, the TDNN map is
\begin{equation}
    \widehat{\mathbf{q}}^{\mathrm{ML}}_c(t_i;\boldsymbol{\phi})
    :=\mathbf{z}^{(L)}_{c,i}
    =\mathcal{F}_{\boldsymbol{\phi}}
    \bigl(\mathbf{u}^{W}_{c,i},v_c\bigr)
    \in\mathbb{R}^{J}.
\label{eq:tdnn_definition}
\end{equation}
Equation~\eqref{eq:tdnn_definition} maps the response history and operating condition to the $J$-channel machine-learning (ML) response generated by the TDNN used in this study. The history vector $\mathbf{u}_{c,i}^{W}\in\mathbb{R}^{JW}$ stacks the delayed simulation responses as $\mathbf{u}_{c,i}^{W}:=\bigl(\mathbf{q}^{\mathrm{s}}_c(t_{i-W+1})^{\mathsf T},\ldots,\mathbf{q}^{\mathrm{s}}_c(t_i)^{\mathsf T}\bigr)^{\mathsf T}$. The augmented network input is $\mathbf{z}^{(0)}_{c,i} := \bigl( (\mathbf{u}^{W}_{c,i})^{\mathsf T}, v_c \bigr)^{\mathsf T} \in\mathbb{R}^{JW+1}$. For layer $\ell=1,\ldots,L$, the weight matrix satisfies $[\mathbf{W}_{\ell}]\in\mathbb{R}^{n_{\ell}\times n_{\ell-1}}$, the bias vector satisfies $\mathbf{b}_{\ell}\in\mathbb{R}^{n_{\ell}}$, and the componentwise activation is $\boldsymbol{\sigma}_{\ell}:\mathbb{R}^{n_{\ell}}\rightarrow\mathbb{R}^{n_{\ell}}$. The preactivation is $\mathbf{a}^{(\ell)}_{c,i}:=[\mathbf{W}_{\ell}]\mathbf{z}^{(\ell-1)}_{c,i}+\mathbf{b}_{\ell}\in\mathbb{R}^{n_{\ell}}$, and the layer output is $\mathbf{z}^{(\ell)}_{c,i}:=\boldsymbol{\sigma}_{\ell}\bigl(\mathbf{a}^{(\ell)}_{c,i}\bigr)\in\mathbb{R}^{n_{\ell}}$. The parameter vector $\boldsymbol{\phi}\in\mathbb{R}^{\sum_{\ell=1}^{L}n_{\ell}(n_{\ell-1}+1)}$ concatenates every entry of the weight matrices and bias vectors.

The delayed window supplies finite temporal memory, while $v_c$ conditions the TDNN response on operating speed. Figure~\ref{fig:tdnn_architecture} illustrates the complete computational graph. Because $\mathbf{u}_{c,i}^{W}$ contains the current simulation response $\mathbf{q}^{\mathrm{s}}_c(t_i)$, $\widehat{\mathbf{q}}^{\mathrm{ML}}_c$ is an unaligned learned representation of an available simulation history, not a one-step-ahead prediction. Replacing the simulation at an unseen condition would require predictive inputs that exclude the target response history and separate validation. Section~\ref{sec:tdnn} defines the development-only fit and identifies the implementation details that remain to be confirmed.

\subsection{Multifidelity Model-Discrepancy Learning} \label{sec:multifidelity_background}

Multifidelity learning combines information sources with different cost, coverage, and proximity to the target physical response so that a structured lower-fidelity trend can be corrected with higher-fidelity evidence \cite{han2012hierarchical,zhang2022physicsInformed,davis2025residual}. For the response quantities and matched roller-rig conditions considered here, we assign the fidelity roles directly to the existing variables:
\begin{equation}
\begin{aligned}
	\mathbf{q}^{\mathrm{e}}_c(t_i)&\in\mathbb{R}^{J}
	&&\text{is the high-fidelity experimental reference},\\[-2pt]
	\mathbf{q}^{\mathrm{s}}_c(t_i)&\in\mathbb{R}^{J}
	&&\text{is the simulation-generated low-fidelity response},\\[-2pt]
	\widehat{\mathbf{q}}^{\mathrm{ML}}_c(t_i;\boldsymbol{\phi})&\in\mathbb{R}^{J}
	&&\text{is the TDNN-derived low-fidelity response}.
\end{aligned}
\label{eq:fidelity_roles}
\end{equation}
The experimental response is high fidelity because it samples the physical bogie at the instrumented locations under the tested conditions. It is a reference, not noise-free ground truth: sensor uncertainty, acquisition noise, preprocessing choices, and finite condition coverage remain. The simulation response is low fidelity because model idealizations, uncertain parameters, and boundary-condition differences can produce systematic discrepancy. This label applies to the selected response quantities and conditions; it does not rank the general accuracy of Simpack or another multibody solver.

The TDNN output remains low fidelity because Eq.~\eqref{eq:tdnn_definition} uses only the simulation history and operating condition; the TDNN does not add experimental information. It supplies a simulation-derived response representation within the correction workflow, while its same-time simulation input makes it a transformed response rather than an independent forecast. After development measurements determine the amplitude alignment, $\widetilde{\mathbf{q}}^{\mathrm{ML}}_c$ is an aligned, high-fidelity-informed low-fidelity baseline. Multifidelity integration begins with this alignment and continues when the residual model learns the remaining discrepancy from the experimental reference. Section~\ref{sec:data_integration} defines the response decomposition, and Section~\ref{sec:residual} introduces the physics-informed correction.

\begin{problem}
Given a limited set of high-fidelity but costly experimental responses and the corresponding lower-fidelity simulation responses, we seek to predict the bogie response that would be measured under an unseen operating condition. The simulation responses provide condition-dependent physical trends, while the experimental responses anchor the prediction to the tested system.
\label{prob:unseen_condition}
\end{problem}

\section{Proposed Multifidelity Simulation--Test Response Correction Method} \label{sec:method}

This section specializes the TDNN and multifidelity concepts introduced in Sections~\ref{sec:tdnn_background} and \ref{sec:multifidelity_background} to the proposed correction workflow. It defines the response decomposition, simulation-derived ML baseline, development-fitted amplitude alignment, physics-informed correction, and complete training and inference procedure.

\subsection{Proposed Method Overview and Response Decomposition} \label{sec:data_integration}

\begin{figure}[htbp]
	\centering
    \includegraphics[width=0.94\linewidth]{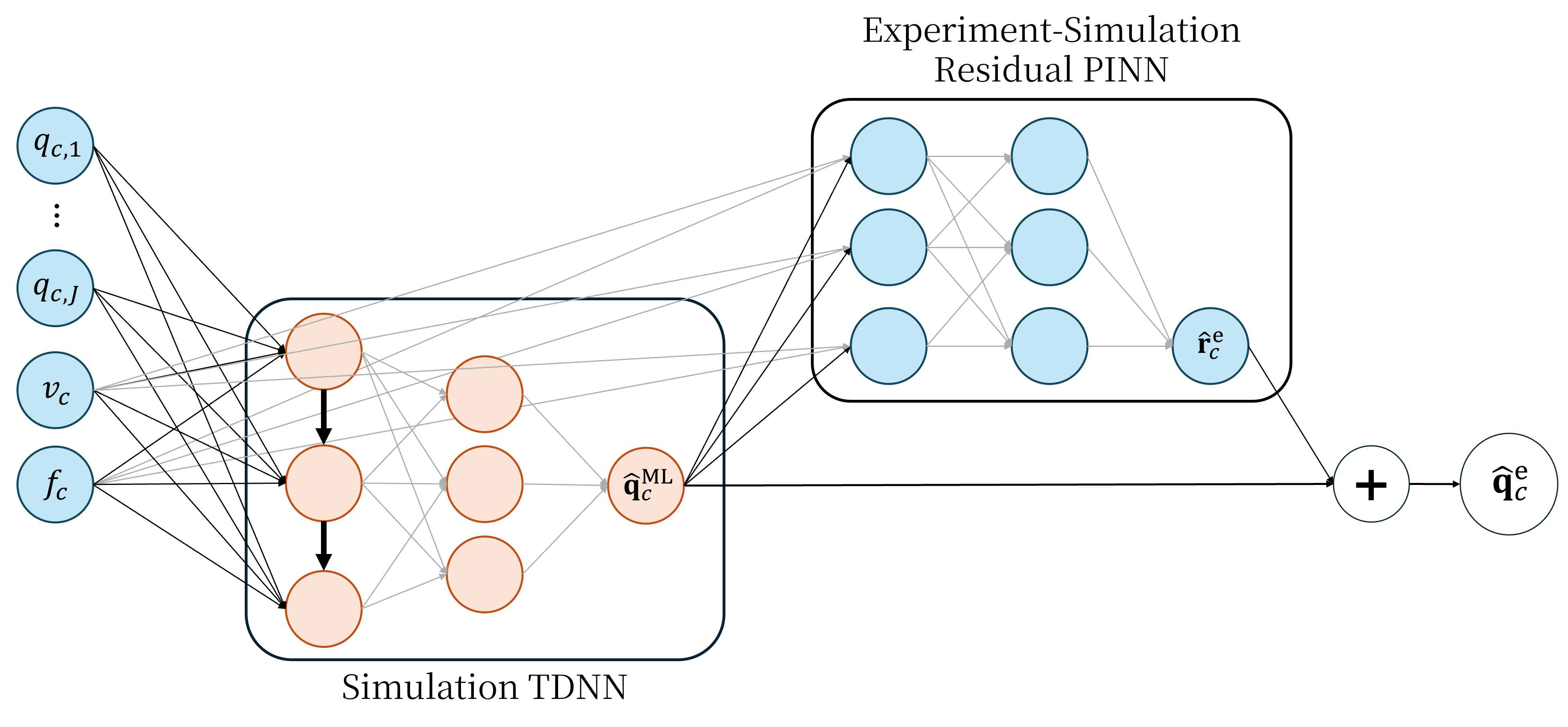}
	\caption{Overview of the proposed TDNN--multifidelity model-discrepancy method. The TDNN maps simulation histories and operating conditions to a condition-dependent baseline. Development measurements align the baseline and define the observed residual. Physics-informed residual learning then adds the learned correction to reconstruct the experimental response.}
	\label{fig:proposed_method_overview}
\end{figure}

We propose a two-stage TDNN--multifidelity model-discrepancy method. Its central idea is to combine experiment-anchored multifidelity learning with a physics-informed constraint on the discrepancy that remains after simulation-based prediction. The TDNN first learns the operating-condition-dependent response trend from simulation histories and produces a baseline informed by simulation. The experimental data then align this baseline in amplitude and define the response component that the simulation does not explain. A physics-informed correction network learns the reproducible part of this experiment--baseline difference and adds it to the baseline to reconstruct the experimental response. Its physics-informed loss constrains the learned correction through a reduced dynamic balance that represents systematic differences in effective inertia, energy dissipation, restoring behavior, and external forcing between the simulation and the physical system. This combined formulation uses the simulation to represent response trends across operating conditions while using the measurements and the effective dynamics to correct systematic model inadequacy. Figure~\ref{fig:proposed_method_overview} illustrates this information flow.

The aligned ML response $\widetilde{\mathbf{q}}^{\mathrm{ML}}_c$ is derived from the simulation response but is not identical to $\mathbf{q}^{\mathrm{s}}_c$. The TDNN represents the simulated history before the development-fitted amplitude transformations are applied. Because this aligned baseline does not reproduce the physical response exactly, we decompose the experimental response and define the observed residual in one relation:
\begin{equation}
\begin{aligned}
	\mathbf{q}^{\mathrm{e}}_c(t_i)
	&=\widetilde{\mathbf{q}}^{\mathrm{ML}}_c(t_i)
	 +\boldsymbol{\delta}_c(t_i)
	 +\boldsymbol{\varepsilon}_c(t_i),\\
	\mathbf{r}^{\mathrm{e}}_c(t_i)
	&:=\mathbf{q}^{\mathrm{e}}_c(t_i)-\widetilde{\mathbf{q}}^{\mathrm{ML}}_c(t_i)
	 =\boldsymbol{\delta}_c(t_i)+\boldsymbol{\varepsilon}_c(t_i).
\end{aligned}
\label{eq:observation_decomposition}
\end{equation}
Here, $\boldsymbol{\delta}_c(t_i),\boldsymbol{\varepsilon}_c(t_i),\mathbf{r}^{\mathrm{e}}_c(t_i)\in\mathbb{R}^{J}$. The vector $\boldsymbol{\delta}_c$ is the latent systematic model discrepancy that remains after the simulation representation and development-fitted amplitude alignment. The vector $\boldsymbol{\varepsilon}_c$ is measurement error, and $\mathbf{r}^{\mathrm{e}}_c$ is their combined observed residual. A deterministic network fitted to one realization cannot identify the two contributions separately; it learns the reproducible component of $\mathbf{r}^{\mathrm{e}}_c$. Sections~\ref{sec:tdnn}--\ref{sec:residual} define the TDNN baseline, alignment, and physics-informed correction in this order.

\subsection{Development Training of the TDNN Baseline} \label{sec:tdnn}

Using the history window and temporal map defined in Eq.~\eqref{eq:tdnn_definition}, we fit the TDNN only to simulation responses from the four development conditions. For $j=1,\ldots,J$, let $s_{q,j}\in\mathbb{R}^{+}$ be the development-derived displacement scale of channel $j$. These conditions provide $N_T:=\sum_{c=1}^{4}(N_c-W+1)$ complete windows, where $N_T\in\mathbb{N}$ and $N_T>0$. The intended development fit is
\begin{equation}
	\boldsymbol{\phi}^{\star}
	\in\operatorname*{arg\,min}_{\boldsymbol{\phi}}
	\frac{1}{N_T}
	\sum_{c=1}^{4}\sum_{i=W-1}^{N_c-1}
	\left\|[\mathbf{S}_{q}]^{-1}
	\bigl(\widehat{\mathbf{q}}^{\mathrm{ML}}_c(t_i;\boldsymbol{\phi})
	-\mathbf{q}^{\mathrm{s}}_c(t_i)\bigr)\right\|_2^2.
\label{eq:tdnn_fit}
\end{equation}
The displacement scale matrix is $[\mathbf{S}_{q}]:=\operatorname{diag}\!\left[\bigl(s_{q,1},\ldots,s_{q,J}\bigr)^{\intercal}\right]\in\mathbb{R}^{J\times J}$. The fitted parameter vector $\boldsymbol{\phi}^{\star}$ is frozen before amplitude alignment. A reproducible implementation must state the objective in Eq.~\eqref{eq:tdnn_fit}, the window length, layer count, layer widths, activation functions, feature order, and inference mode.

\subsection{Development-Fitted Amplitude Alignment} \label{sec:tdnn_align}

We next apply one development-fitted amplitude factor to the frozen ML response. The aligned ML response is
\begin{equation}
\begin{aligned}
\widetilde{\mathbf{q}}^{\mathrm{ML}}_c(t_i)
&:=\rho^{\star}\widehat{\mathbf{q}}^{\mathrm{ML}}_c
(t_i;\boldsymbol{\phi}^{\star})
\in\mathbb{R}^{J},\\
\rho^{\star}
&:=\left(
\frac{
\displaystyle\sum_{c=1}^{4}
\sum_{i=W-1}^{N_c-1}
\left\|\mathbf{q}^{\mathrm{e}}_c(t_i)\right\|_2^2
}{
\displaystyle\sum_{c=1}^{4}
\sum_{i=W-1}^{N_c-1}
\left\|\widehat{\mathbf{q}}^{\mathrm{ML}}_c
(t_i;\boldsymbol{\phi}^{\star})\right\|_2^2
}
\right)^{1/2}.
\end{aligned}
\label{eq:alignment_fit}
\end{equation}

The amplitude-factor definition in Eq.~\eqref{eq:alignment_fit} requires a nonzero denominator and gives
$\rho^{\star}\in\mathbb{R}$ with $\rho^{\star}\geq 0$.
We estimate $\rho^{\star}$ from development records and then freeze it.

\subsection{Physics-Informed Residual Correction} \label{sec:residual}

Let $p\in\mathbb{N}$, $p>0$, denote the reduced-coordinate dimension. The residual network $\mathcal{G}_{\boldsymbol{\theta}}: \mathbb{R}\times\mathbb{R}^{J}\times\mathbb{R}^{3} \rightarrow\mathbb{R}^{p}$ maps time, the aligned ML response, and the PINN input vector $\boldsymbol{\xi}_c(t_i)$ to a reduced-coordinate correction. Its parameter vector satisfies $\boldsymbol{\theta}\in\mathbb{R}^{n_{\theta}}$, where $n_{\theta}\in\mathbb{N}$ and $n_{\theta}>0$. The final corrected response is
\begin{equation}
	\widehat{\mathbf{q}}_c(t_i;\boldsymbol{\theta})
	:=\widetilde{\mathbf{q}}^{\mathrm{ML}}_c(t_i)+\boldsymbol{\Delta}_{\boldsymbol{\theta},c}(t_i)
	\in\mathbb{R}^{J}.
\label{eq:multifidelity}
\end{equation}
The reduced-coordinate network output is $\widehat{\boldsymbol{\delta}}^{y}_{\boldsymbol{\theta},c}(t_i):=\mathcal{G}_{\boldsymbol{\theta}}\bigl(t_i,\widetilde{\mathbf{q}}^{\mathrm{ML}}_c(t_i),\boldsymbol{\xi}_c(t_i)\bigr)\in\mathbb{R}^{p}$. Its sensor-space projection is $\boldsymbol{\Delta}_{\boldsymbol{\theta},c}(t_i):=[\mathbf{H}_q]\widehat{\boldsymbol{\delta}}^{y}_{\boldsymbol{\theta},c}(t_i)\in\mathbb{R}^{J}$. The observation matrix $[\mathbf{H}_q]\in\mathbb{R}^{J\times p}$ maps the reduced-coordinate correction to the measured channels. This matrix must be established from a verified sensor-to-coordinate map before the reduced-coordinate correction is interpreted physically. The special case $p=J$ and $[\mathbf{H}_q]=[\mathbf{I}_J]$ is valid only if the four measured outputs themselves form a mechanically consistent reduced-coordinate vector, where $[\mathbf{I}_J]$ denotes the $J\times J$ identity matrix.

To give the correction a mechanical interpretation, let $\mathbf{y}_{s,c}(t_i)\in\mathbb{R}^{p}$ denote the nominal reduced coordinate, and let $\mathbf{y}_{r,c}(t_i)\in\mathbb{R}^{p}$ denote the physical reduced coordinate. We require $\widetilde{\mathbf{q}}^{\mathrm{ML}}_c(t_i)=[\mathbf{H}_q]\mathbf{y}_{s,c}(t_i)$ and $\boldsymbol{\delta}_c(t_i)=[\mathbf{H}_q]\boldsymbol{\delta}^{y}_c(t_i)$. Because amplitude alignment is applied at the sensor-response interface, $\mathbf{y}_{s,c}$ is an effective reduced coordinate chosen to reproduce $\widetilde{\mathbf{q}}^{\mathrm{ML}}_c$, and $\mathbf{f}_{s,c}$ is its corresponding effective generalized force. Neither quantity is necessarily an unmodified state or force history exported by the multibody solver. The nominal and physical reduced systems satisfy
\begin{equation}
\begin{aligned}
	[\mathbf{M}_{s}]\ddot{\mathbf{y}}_{s,c}
	+[\mathbf{C}_{s}]\dot{\mathbf{y}}_{s,c}
	+[\mathbf{K}_{s}]\mathbf{y}_{s,c}
	&=\mathbf{f}_{s,c},\\
	[\mathbf{M}_{r}]\ddot{\mathbf{y}}_{r,c}
	+[\mathbf{C}_{r}]\dot{\mathbf{y}}_{r,c}
	+[\mathbf{K}_{r}]\mathbf{y}_{r,c}
	&=\mathbf{f}_{r,c}.
\end{aligned}
\label{eq:nominal_real_dynamics}
\end{equation}
The effective matrix discrepancies are $[\Delta\mathbf{M}]:=[\mathbf{M}_{r}]-[\mathbf{M}_{s}]$, $[\Delta\mathbf{C}]:=[\mathbf{C}_{r}]-[\mathbf{C}_{s}]$, and $[\Delta\mathbf{K}]:=[\mathbf{K}_{r}]-[\mathbf{K}_{s}]$. The force discrepancy is $\Delta\mathbf{f}_c:=\mathbf{f}_{r,c}-\mathbf{f}_{s,c}$, and the coordinate discrepancy is $\boldsymbol{\delta}^{y}_c:=\mathbf{y}_{r,c}-\mathbf{y}_{s,c}$. All mass, damping, stiffness, and parameter-discrepancy matrices in Eq.~\eqref{eq:nominal_real_dynamics} belong to $\mathbb{R}^{p\times p}$. The vectors $\mathbf{f}_{s,c}$, $\mathbf{f}_{r,c}$, $\Delta\mathbf{f}_c$, and $\boldsymbol{\delta}^{y}_c$ belong to $\mathbb{R}^{p}$. Replacing the exact coordinate discrepancy by the network output gives the compact effective-dynamics residual below; all terms are evaluated at the same time, which we omit for readability.
\begin{equation}
\begin{aligned}
	\mathbf{g}_{\boldsymbol{\theta},c}
	&:=\bigl([\mathbf{M}_{s}]+[\Delta\mathbf{M}]\bigr)
	\bigl(\ddot{\mathbf{y}}_{s,c}+\ddot{\widehat{\boldsymbol{\delta}}}^{y}_{\boldsymbol{\theta},c}\bigr)
	+\bigl([\mathbf{C}_{s}]+[\Delta\mathbf{C}]\bigr)
	\bigl(\dot{\mathbf{y}}_{s,c}+\dot{\widehat{\boldsymbol{\delta}}}^{y}_{\boldsymbol{\theta},c}\bigr)\\
	&\quad+\bigl([\mathbf{K}_{s}]+[\Delta\mathbf{K}]\bigr)
	\bigl(\mathbf{y}_{s,c}+\widehat{\boldsymbol{\delta}}^{y}_{\boldsymbol{\theta},c}\bigr)
	-\bigl(\mathbf{f}_{s,c}+\Delta\mathbf{f}_c\bigr).
\end{aligned}
\label{eq:physics_residual}
\end{equation}
The residual satisfies $\mathbf{g}_{\boldsymbol{\theta},c}(t_i)\in\mathbb{R}^{p}$, and the ideal correction makes $\mathbf{g}_{\boldsymbol{\theta},c}=\mathbf{0}$. The matrices represent effective differences between reduced systems, not identified errors in individual suspension components. A reproducible implementation must prescribe or estimate $[\Delta\mathbf{M}]$, $[\Delta\mathbf{C}]$, $[\Delta\mathbf{K}]$, and a constrained finite-dimensional representation of $\Delta\mathbf{f}_c$ using development data. Any identified matrices require an admissible parameterization: in particular, the effective mass matrix must remain symmetric positive definite, while damping and stiffness constraints must be consistent with the selected reduced linearization. The present analysis does not identify these matrices. Equation~\eqref{eq:physics_residual} is therefore a corrective-source regularizer, not an exact reparameterization of the complete nonlinear wheel--roller multibody model.

We evaluate time derivatives on the uniform grid with centered differences. The channel-2 acceleration relation is
\begin{equation}
\widehat a_{c,2}(t_i;\boldsymbol{\theta})
:= \gamma_q D_{\Delta t}^{2}\widehat q_{c,2}(t_i;\boldsymbol{\theta}).
\label{eq:predicted_acceleration}
\end{equation}
For a scalar history, let $z_i:=z(t_i)\in\mathbb{R}$. The first centered-difference operator is $D_{\Delta t}z_i:=(z_{i+1}-z_{i-1})/(2\Delta t)$, and the second centered-difference operator is $D_{\Delta t}^{2}z_i:=(z_{i+1}-2z_i+z_{i-1})/(\Delta t)^2$. Both operators act componentwise on vector histories. The conversion factor is $\gamma_q:=10^{-3}\ \si{\metre\per\milli\metre}$. The predicted and measured accelerations, $\widehat a_{c,2}(t_i;\boldsymbol{\theta})$ and $a^{\mathrm e}_{c,2}(t_i)$, are real scalars. All derivative-based terms exclude boundary samples without the required neighbors. The same point, axis, direction, reference frame, bandwidth, and kinematic quantity must be verified before Eq.~\eqref{eq:predicted_acceleration} is interpreted as a physical equality.

For each loss label $\nu$, let $i^-_{\nu,c},i^+_{\nu,c}\in\mathbb{N}$ denote the first and last valid sequential sample indices in development condition $c$. The label $\nu$ is $\mathrm{res}$, $\mathrm{smooth}$, $a$, $a_{\mathrm{kin}}$, $a_{\mathrm{cons}}$, or $\mathrm{phys}$, and the corresponding sample count is \[N_{\nu}:=\sum_{c=1}^{4}\left(i^+_{\nu,c}-i^-_{\nu,c}+1\right).\] The residual bounds retain aligned samples, the smoothness bounds remove centered-difference boundary samples, the acceleration bounds include valid channel-2 pairs, and the physics bounds retain samples with every reduced-state and force quantity required by Eq.~\eqref{eq:physics_residual}. All corresponding sample counts are required to be positive.

The matrices $[\mathbf S_q],[\mathbf S_\kappa]\in\mathbb{R}^{J\times J}$ and $[\mathbf S_g]\in\mathbb{R}^{p\times p}$ are positive diagonal scaling matrices. The scalar $s_{a,2}\in\mathbb{R}^{+}$ has acceleration units. All scales are fixed from development data. We suppress the common time argument $t_i$ and the already defined parameter dependence of $\widehat a_{c,2}$ in the loss expressions. The training objective is written using four loss groups, with the acceleration group combining three weighted channel-2 penalties:
\begin{subequations}\label{eq:loss_components}
\begin{align}
\boldsymbol{\theta}^{\star}
&\in \operatorname*{arg\,min}_{\boldsymbol{\theta}}
\mathcal{L}(\boldsymbol{\theta}),
\label{eq:loss}
\\
\mathcal{L}(\boldsymbol{\theta})
&:= \lambda_{\mathrm{res}}\mathcal{L}_{\mathrm{res}}
+ \lambda_{\mathrm{smooth}}\mathcal{L}_{\mathrm{smooth}}
+ \mathcal{L}_{\mathrm{acc}}
+ \lambda_{\mathrm{phys}}\mathcal{L}_{\mathrm{phys}},
\label{eq:total_loss}
\\
\mathcal{L}_{\mathrm{res}}
&:= \frac{1}{N_{\mathrm{res}}}\sum_{c=1}^{4}\sum_{i=i^-_{\mathrm{res},c}}^{i^+_{\mathrm{res},c}}
\left\|[\mathbf S_q]^{-1}\left(\boldsymbol{\Delta}_{\boldsymbol{\theta},c}
- \mathbf r^{\mathrm e}_{c} \right) \right\|_2^2,
\\
\mathcal{L}_{\mathrm{smooth}}
&:= \frac{1}{N_{\mathrm{smooth}}}\sum_{c=1}^{4}\sum_{i=i^-_{\mathrm{smooth},c}}^{i^+_{\mathrm{smooth},c}}\left\|
[\mathbf S_\kappa]^{-1} D_{\Delta t}^{2} \boldsymbol{\Delta}_{\boldsymbol{\theta},c}
\right\|_2^2,
\\
\mathcal{L}_{\mathrm{acc}}
&:= \frac{\lambda_{a,2}}{N_a}\sum_{c=1}^{4}\sum_{i=i^-_{a,c}}^{i^+_{a,c}}
\left(\frac{\left[\mathcal B_c\left(\widehat a^{\mathrm h}_{c,2}
- a^{\mathrm e}_{c,2} \right) \right]_i}{s_{a,2}}\right)^2
\nonumber\\
&\quad+
\frac{\lambda_{a,2}^{\mathrm{kin}}}{N_{a,\mathrm{kin}}}
\sum_{c=1}^{4}\sum_{i=i^-_{a,\mathrm{kin},c}}^{i^+_{a,\mathrm{kin},c}}
\left(\frac{\left[\mathcal B_c\left(\widehat a_{c,2}
- a^{\mathrm e}_{c,2}\right)\right]_i}{s_{a,2}}\right)^2
\nonumber\\
&\quad+
\frac{\lambda_{a,2}^{\mathrm{cons}}}{N_{a,\mathrm{cons}}}
\sum_{c=1}^{4}\sum_{i=i^-_{a,\mathrm{cons},c}}^{i^+_{a,\mathrm{cons},c}}
\left(\frac{\left[\mathcal B_c\left(\widehat a^{\mathrm h}_{c,2}
- \widehat a_{c,2} \right) \right]_i}{s_{a,2}}\right)^2,
\label{eq:acceleration_loss}
\\
\mathcal{L}_{\mathrm{phys}}
&:=\frac{1}{N_{\mathrm{phys}}}\sum_{c=1}^{4}\sum_{i=i^-_{\mathrm{phys},c}}^{i^+_{\mathrm{phys},c}}
\left\|[\mathbf S_g]^{-1}\mathbf g_{\boldsymbol{\theta},c}\right\|_2^2.
\end{align}
\end{subequations}

All loss terms and $\mathcal{L}(\boldsymbol{\theta})$ are nonnegative real scalars. The weights $\lambda_{\mathrm{res}}$, $\lambda_{\mathrm{smooth}}$, $\lambda_{a,2}$, $\lambda_{a,2}^{\mathrm{kin}}$, $\lambda_{a,2}^{\mathrm{cons}}$, and $\lambda_{\mathrm{phys}}$ are nonnegative real scalars. As defined in Eq.~\eqref{eq:predicted_acceleration}, $\widehat a_{c,2}$ is the channel-2 acceleration derived from the corrected displacement, whereas $\widehat a^{\mathrm h}_{c,2}$ denotes the corresponding acceleration-head prediction.

The combined acceleration loss $\mathcal{L}_{\mathrm{acc}}$ contains three weighted penalties: direct matching between the acceleration-head prediction and the measured acceleration, kinematic matching between the displacement-derived and measured accelerations, and consistency between the acceleration-head and displacement-derived accelerations. The operator $\mathcal B_c$ denotes the condition-dependent coherent-band projection applied to these terms. A common passband of 3.5--4.5~Hz is used for all development conditions; see~\ref{app:channel_diagnostic} for the corresponding channel-diagnostic results. The residual term fits the measured correction, the smoothness term suppresses rapidly varying correction curvature, and the physics term penalizes the effective dynamic imbalance. The individual  contributions of the three constituent acceleration penalties are not isolated in the present study and require controlled ablation.

\subsection{Overall Training and Inference Workflow} \label{sec:overall_algorithm}

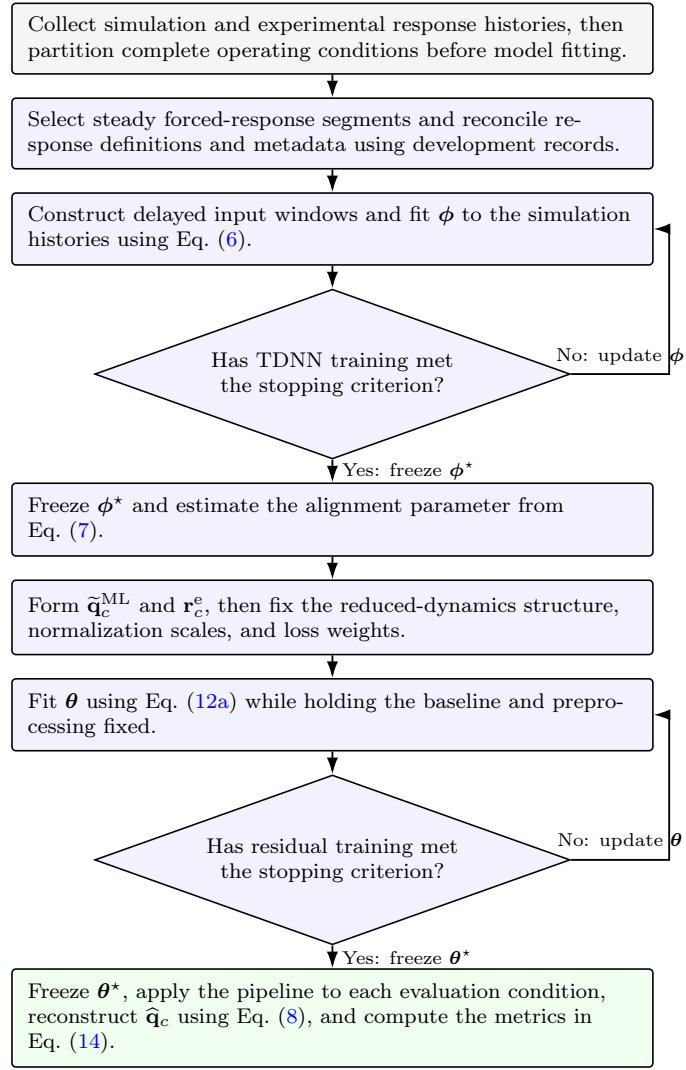
\begin{figure}[H]
	\centering
	\begin{tikzpicture}[
		node distance=3.0mm,
		flow/.style={
			rectangle,
			rounded corners=1.5pt,
			draw=black,
			line width=0.6pt,
			text width=0.52\linewidth,
			align=left,
			inner xsep=6pt,
			inner ysep=5pt,
			font=\footnotesize
		},
		input/.style={flow,fill=black!4},
		development/.style={flow,fill=blue!5},
		evaluation/.style={flow,fill=green!6},
		decision/.style={
			diamond,
			aspect=2.8,
			draw=black,
			line width=0.6pt,
			fill=blue!5,
			text width=0.28\linewidth,
			align=center,
			inner sep=1.2pt,
			font=\footnotesize
		},
		arrow/.style={-{Latex[length=2.2mm,width=1.5mm]},line width=0.7pt}
	]
		\node[input] (input) {
			Collect simulation and experimental response histories, then partition complete operating conditions before model fitting.
		};
		\node[development,below=of input] (prepare) {
			Select steady forced-response segments and reconcile response definitions and metadata using development records.
		};
		\node[development,below=of prepare] (tdnn) {
			Construct delayed input windows and fit $\boldsymbol{\phi}$ to the simulation histories using Eq.~\eqref{eq:tdnn_fit}.
		};
		\node[decision,below=of tdnn] (tdnnstop) {
			Has TDNN training met the stopping criterion?
		};
		\node[development,below=of tdnnstop] (align) {
			Freeze $\boldsymbol{\phi}^{\star}$ and estimate the alignment parameter from Eq.~\eqref{eq:alignment_fit}.
		};
		\node[development,below=of align] (residual) {
			Form $\widetilde{\mathbf{q}}^{\mathrm{ML}}_c$ and $\mathbf{r}^{\mathrm{e}}_c$, then fix the reduced-dynamics structure, normalization scales, and loss weights.
		};
		\node[development,below=of residual] (train) {
			Fit $\boldsymbol{\theta}$ using Eq.~\eqref{eq:loss} while holding the baseline and preprocessing fixed.
		};
		\node[decision,below=of train] (resstop) {
			Has residual training met the stopping criterion?
		};
		\node[evaluation,below=of resstop] (evaluate) {
			Freeze $\boldsymbol{\theta}^{\star}$, apply the pipeline to each evaluation condition, reconstruct $\widehat{\mathbf{q}}_c$ using Eq.~\eqref{eq:multifidelity}, and compute the metrics in Eq.~\eqref{eq:metrics}.
		};
		\draw[arrow] (input) -- (prepare);
		\draw[arrow] (prepare) -- (tdnn);
		\draw[arrow] (tdnn) -- (tdnnstop);
		\draw[arrow] (tdnnstop) -- node[right,font=\scriptsize]{Yes: freeze $\boldsymbol{\phi}^{\star}$} (align);
		\draw[arrow] (tdnnstop.east) -- ++(13mm,0)
			node[midway,above,font=\scriptsize]{No: update $\boldsymbol{\phi}$}
			|- (tdnn.east);
		\draw[arrow] (align) -- (residual);
		\draw[arrow] (residual) -- (train);
		\draw[arrow] (train) -- (resstop);
		\draw[arrow] (resstop) -- node[right,font=\scriptsize]{Yes: freeze $\boldsymbol{\theta}^{\star}$} (evaluate);
		\draw[arrow] (resstop.east) -- ++(13mm,0)
			node[midway,above,font=\scriptsize]{No: update $\boldsymbol{\theta}$}
			|- (train.east);
	\end{tikzpicture}
	\caption{Training and inference workflow for the proposed TDNN--multifidelity model-discrepancy method. The two decision nodes return optimization to the corresponding training step until its prespecified stopping criterion is satisfied. Blue nodes denote development operations, the gray node denotes data preparation, and the green node denotes evaluation with all target-informed quantities frozen.}
	\label{fig:overall_workflow}
\end{figure}

Figure~\ref{fig:overall_workflow} illustrates the method in the order in which information becomes available. Every target-informed operation uses development data, and all fitted transformations and networks remain fixed during evaluation. The interface is solver independent because it requires compatible response histories rather than a Simpack-specific state representation.

\section{Results and Discussion} \label{sec:results}

The multibody simulations were performed on a workstation running Microsoft Windows 11 Pro 25H2, equipped with an Intel Core Ultra 5 225 processor with 10 physical cores and 32 GB of memory; the simulations used SIMULIA Simpack 2025x.2. Data preprocessing and learning-model training were performed on a workstation running Microsoft Windows 11 Pro 23H2, equipped with an Intel Core Ultra 5 225 processor with 10 physical cores and 32 GB of memory, and an NVIDIA GeForce RTX 3060 with 12 GB of memory. The learning environment used Python 3.11.16, PyTorch 2.9.1+cu128, PyTorch CUDA version: 12.8, NumPy 2.4.6, SciPy 1.17.1, and pandas 3.0.5.

We present three evidence layers. The multibody-model assessment establishes numerical and mechanical consistency within its defined scope. The experiment-only PINN and TDNN-assisted residual model provide learning-based reference cases. The proposed discrepancy-correction analysis evaluates the reconstructed response associated with the revised residual formulation. These layers answer different questions and are not interchangeable.

We distinguish three response-reconstruction cases at the start of the results. The experiment-only PINN receives no simulation response. The TDNN-assisted residual model first represents the simulation response and then adds a learned residual. The proposed discrepancy-correction formulation retains the aligned TDNN baseline and constrains its learned correction using the response decomposition in Eq.~\eqref{eq:observation_decomposition} and effective-physics regularization. Table~\ref{tab:case_definition} defines the information available for each case.

For all three learning-based cases, model training and cross-validation use only the 300--360~km/h development conditions. The 385~km/h condition is reserved exclusively for extrapolative evaluation and is not used for parameter fitting, hyperparameter selection, scaler fitting, checkpoint selection, or any other target-informed training operation.

\begin{table}[htbp]
	\centering
	\caption{Inputs and reconstructed outputs of the learning-based comparison cases.}
	\label{tab:case_definition}
	\begin{tabular}{>{\raggedright\arraybackslash}p{0.32\textwidth}>{\raggedright\arraybackslash}p{0.26\textwidth}>{\raggedright\arraybackslash}p{0.32\textwidth}}
		\hline
        \noalign{\vskip 0.3ex}
		Case & Simulation input & Predicted quantity \\
		\hline
		Experiment-only PINN & None & Experimental response \\
        \\
		TDNN-assisted residual model & TDNN representation of the simulation response & TDNN baseline plus an additive learned residual \\
        \\
		Proposed discrepancy-correction formulation & Aligned ML baseline and condition input & Aligned ML baseline plus a physics-constrained correction \\
		\hline
	\end{tabular}
\end{table}

To define a common target condition across these layers, we use the intended evaluation condition $c=5$. Let $\widehat q^{m}_{5,j}(t_i)$ denote the prediction from one compared model $m$ and let $q^{\mathrm{e}}_{5,j}(t_i)$ denote the paired measurement, where $i=0,\ldots,N_5-1$. Define the experimental sample mean and response range for channel $j$ as
\begin{equation}
\overline q^{\mathrm{e}}_{5,j}
:=\frac{1}{N_5}\sum_{i=0}^{N_5-1}q^{\mathrm{e}}_{5,j}(t_i),
\qquad
\Delta q^{\mathrm{e}}_{5,j}
:=\max_i q^{\mathrm{e}}_{5,j}(t_i)-\min_i q^{\mathrm{e}}_{5,j}(t_i).
\label{eq:evaluation_range}
\end{equation}

The channelwise metrics are
\begin{equation}
\begin{aligned}
R_{j}^{2}(m)&:=1-\frac{\sum_{i=0}^{N_5-1}\bigl(q^{\mathrm{e}}_{5,j}(t_i)
-\widehat q^{m}_{5,j}(t_i)\bigr)^2}{\sum_{i=0}^{N_5-1}\bigl(q^{\mathrm{e}}_{5,j}(t_i)
-\overline q^{\mathrm{e}}_{5,j}\bigr)^2},
\\
\operatorname{NRMSE}_{j}(m)
&:=\frac{\left[\frac{1}{N_5}\sum_{i=0}^{N_5-1}\bigl(q^{\mathrm{e}}_{5,j}(t_i)-\widehat q^{m}_{5,j}(t_i)\bigr)^2\right]^{1/2}}{\Delta q^{\mathrm{e}}_{5,j}}\times 100\%,
\\
\operatorname{NMAE}_{j}(m)
&:=\frac{\frac{1}{N_5}\sum_{i=0}^{N_5-1}\left|q^{\mathrm{e}}_{5,j}(t_i)
-\widehat q^{m}_{5,j}(t_i)\right|}{\Delta q^{\mathrm{e}}_{5,j}}
\times 100\%.
\end{aligned}
\label{eq:metrics}
\end{equation}

The coefficient $R_j^2(m)\in\mathbb{R}$, while $\operatorname{NRMSE}_j(m)$ and $\operatorname{NMAE}_j(m)$ are nonnegative percentage-valued scalars. We evaluate $R_j^2$ only when the experimental total sum of squares in its denominator is nonzero, and the normalized error metrics only when $\Delta q^{\mathrm{e}}_{5,j}>0$. The normalization range is determined only from the measured 385~km/h response of each channel and is therefore common to all compared models. The mean metric is the arithmetic average of its four channel values. The model index $m$ denotes the experiment-only PINN, TDNN-assisted residual model, or proposed discrepancy-correction formulation.

\subsection{Simulation Model Consistency} \label{sec:simulation_results}

\begin{figure}[htbp]
    \centering

    \begin{subfigure}[b]{0.38\linewidth}
        \centering
        \includegraphics[width=\linewidth]
        {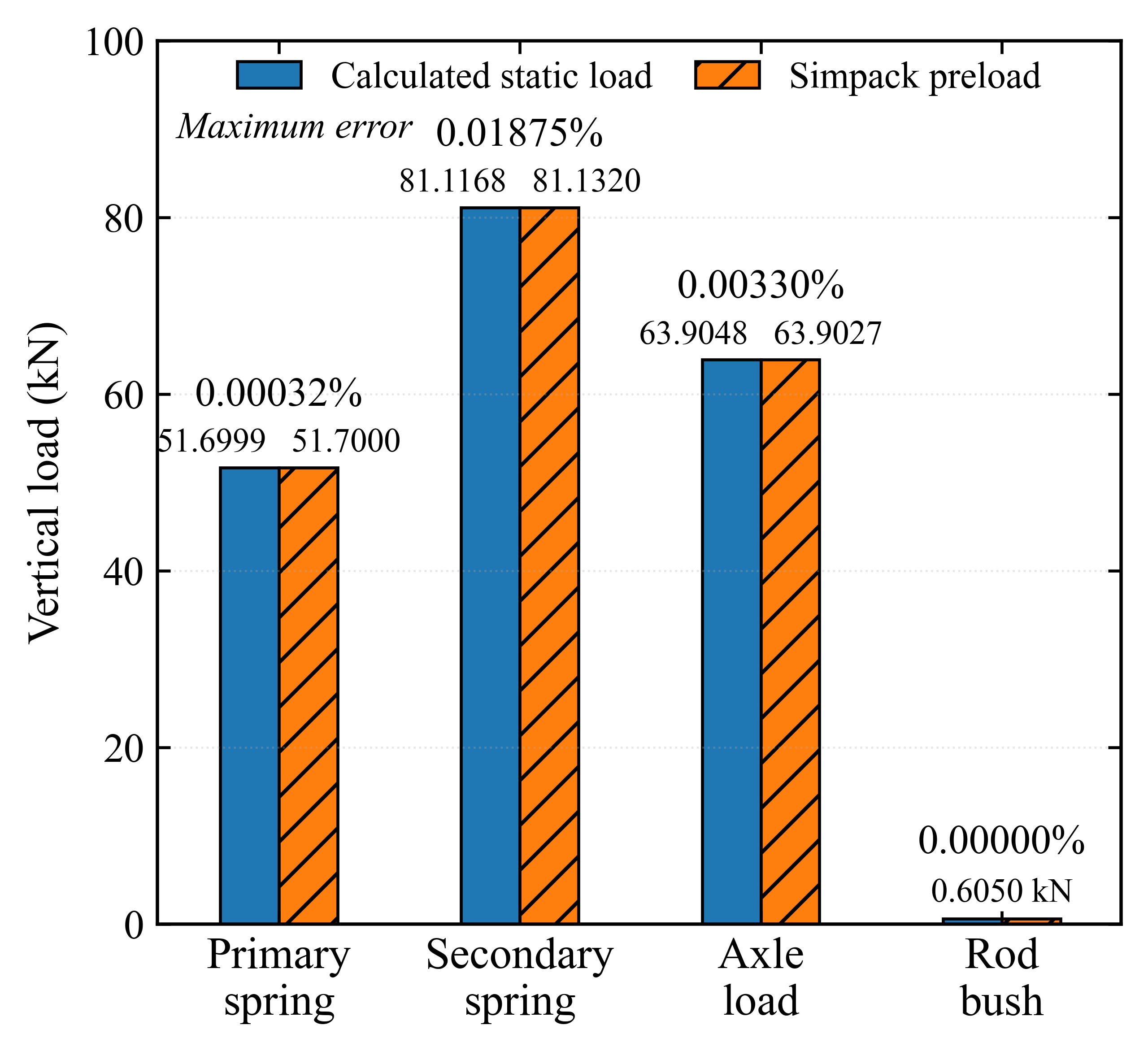}
        \caption{}
        \label{fig:model_preload_comparison}
    \end{subfigure}
    \hfill
    \begin{subfigure}[b]{0.59\linewidth}
        \centering
        \includegraphics[width=\linewidth]
        {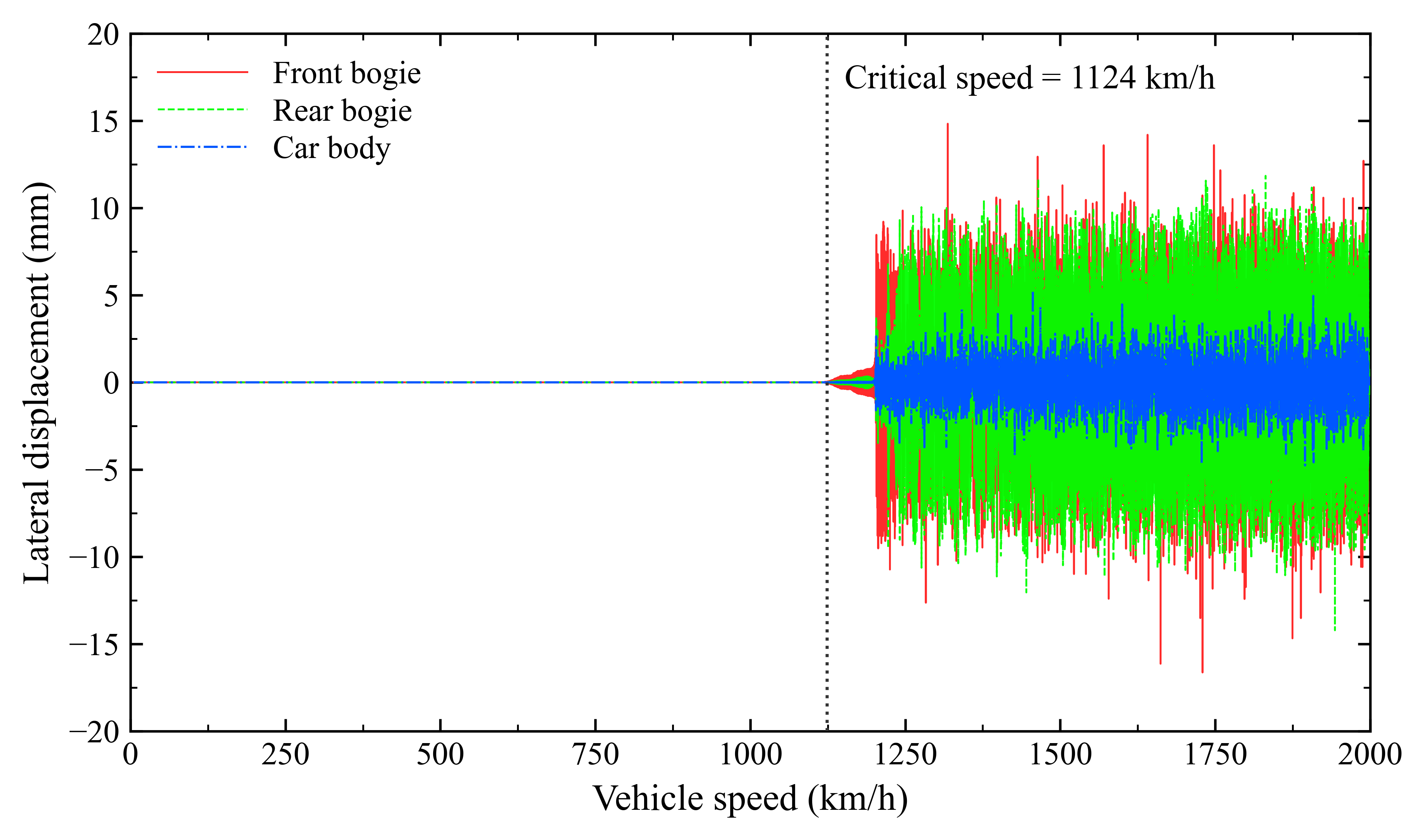}
        \caption{}
        \label{fig:model_critical_speed}
    \end{subfigure}

    \caption{
    Internal-consistency assessment of the full-vehicle model:
    (a) comparison of independently calculated static-equilibrium loads
    and the corresponding Simpack preload-solution quantities for the
    primary spring, secondary spring, axle load, and rod-bush load, where
    each pair represents the case with the largest absolute relative error
    within the corresponding quantity group;
    (b) lateral-displacement responses of the front bogie, rear bogie,
    and car body during the nonlinear speed sweep. The vertical dotted
    line indicates the identified critical speed of
    \SI{1124}{\kilo\metre\per\hour}.
    }
    \label{fig:model_internal_plots}
\end{figure}

All monitored preload force elements converge, and the maximum residual constraint acceleration reaches \SI{0.0189}{\metre\per\second\squared}, below the \SI{0.1}{\metre\per\second\squared} screening threshold. Selected suspension, rod-bush, and axle-load quantities agree with independently calculated static-equilibrium values within a 5\% screening tolerance. These findings support only static assembly and load-transfer consistency. We also process selected running-safety quantities with EN~14363-based procedures and compare them descriptively with available Vampire outputs. Across the three analyzed curve radii, lateral force, both wheel-unloading ratios, and the left-wheel derailment quotient share the same rank order; the right-wheel derailment quotient shows only partial rank agreement. The nonlinear speed sweep identifies a transition from negligible lateral response to sustained lateral oscillation at approximately \SI{1124}{\kilo\metre\per\hour}, which is taken as the nonlinear critical speed of the Simpack model. The available cross-tool critical-speed values are excluded from direct validation because the corresponding model configurations are not matched. Figure~\ref{fig:model_internal_plots} presents the model-consistency results, and Table~\ref{tab:model_assessment} states the supported interpretation of each assessment item.

\begin{table}[H]
	\centering
	\caption{Scope and interpretation of the simulation-model consistency assessment.}
	\label{tab:model_assessment}
	\begin{tabular}{>{\raggedright\arraybackslash}p{0.20\textwidth}>{\raggedright\arraybackslash}p{0.32\textwidth}p{0.37\textwidth}}
		\hline
        \noalign{\vskip 0.3ex}
		Assessment item & Evidence examined & Supported interpretation \\
		\hline
		Preload and static equilibrium & Residual-acceleration screening and independent load balance within the 5\% screening tolerance & Consistent static assembly and load transfer \\
		\hline
		Running-safety post-processing & Descriptive rank comparison across three curve radii & Limited cross-code consistency; no compliance or magnitude validation \\
		\hline
		Nonlinear critical speed &
        Lateral-displacement response during the nonlinear speed sweep and transition at \SI{1124}{\kilo\metre\per\hour} &
        Identified hunting-stability transition of the Simpack model; cross-tool values not directly comparable \\
		\hline
	\end{tabular}
\end{table}

These assessments provide limited internal-consistency evidence for the full-vehicle model within the examined numerical cases. They do not show that its predictions can replace repeated development tests, nor do they quantify the accuracy of the four roller-rig histories.

\subsection{Experiment-Only PINN and Simulation-Informed Comparison} \label{sec:pinn_results}

The experiment-only PINN predicts the four experimental displacement responses directly, without using the simulation response or the TDNN baseline. Its training objective is
\begin{equation}
\begin{gathered}
\boldsymbol{\theta}^{\star}_\mathrm{PINN}
\in\operatorname*{arg\,min}_{\boldsymbol{\theta}_\mathrm{PINN}}\mathcal{L}_{\mathrm{PINN}}(\boldsymbol{\theta}_\mathrm{PINN}),\\
\mathcal{L}_{\mathrm{PINN}}(\boldsymbol{\theta}_\mathrm{PINN})
=
\mathcal{L}_{\mathrm{data}}
+
\lambda^\mathrm{PINN}_{\mathrm{smooth}}
\mathcal{L}^\mathrm{PINN}_{\mathrm{smooth}}
+
\lambda_{\mathrm{SDOF}}
\mathcal{L}_{\mathrm{SDOF}},
\end{gathered}
\label{eq:experiment_only_loss}
\end{equation}
where $\mathcal{L}_{\mathrm{data}}$ is the mean squared error between the predicted and measured displacements standardized using the development data. The smoothness term penalizes the second discrete difference of the predicted displacement within each operating condition.

Unlike the proposed discrepancy-correction formulation, this model does not use the residual-matching loss $\mathcal{L}_{\mathrm{res}}$, because it predicts the experimental response directly rather than a correction to $\widetilde{\mathbf{q}}^{\mathrm{ML}}_c$. The three acceleration-related penalties corresponding to the constituent terms of $\mathcal{L}_{\mathrm{acc}}$ were considered during development-only model selection but were omitted because they did not improve held-out displacement accuracy. This difference may partly reflect the fact that the experiment-only PINN must reconstruct the complete displacement history directly, making the additional acceleration constraint more likely to compete with the primary displacement-fitting objective. The reduced dynamic-residual loss $\mathcal{L}_{\mathrm{phys}}$ in Eq.~\eqref{eq:loss_components} is replaced by a channelwise effective single-degree-of-freedom (SDOF) regularization,
\begin{equation}
g^{\mathrm{PINN}}_{c,j}(t_i)
:=
\gamma_q
\left(
\ddot{\widehat q}^{\mathrm{PINN}}_{c,j}
+
2\zeta_j\omega_{n,j}
\dot{\widehat q}^{\mathrm{PINN}}_{c,j}
+
\omega_{n,j}^2
\widehat q^{\mathrm{PINN}}_{c,j}
\right)
-
\left[
A_j(v_c)\sin(\omega_c t_i)
+
B_j(v_c)\cos(\omega_c t_i)
\right],
\label{eq:experiment_only_sdof}
\end{equation}
with
\begin{equation}
\mathcal{L}_{\mathrm{SDOF}}:=\operatorname{mean}_{c,j,i}
\left(\frac{g^{\mathrm{PINN}}_{c,j}(t_i)}{s_{\mathrm{SDOF},j}}\right)^2 .
\end{equation}
Here, $\omega_{\mathrm{c}}=2\pi f_{\mathrm{c}}$ with $f_{\mathrm{c}}=2~\mathrm{Hz}$, while the natural frequency $\omega_{n,j}$ and damping ratio $\zeta_j$ are trainable for each response channel. The harmonic forcing coefficients are affine functions of a normalized form of $v_c$. Thus, the experiment-only physics term provides a generic dynamic regularization and does not use the simulation--experiment discrepancy parameters of Eq.~\eqref{eq:physics_residual}.

The reported hyperparameters were selected through development-only leave-one-speed-out cross-validation. The selected PINN uses hidden layers of 96, 96, and 64 neurons with hyperbolic-tangent activations. Adam is used with a learning rate of $2\times10^{-3}$ and a weight decay of $10^{-6}$, while the smoothness and SDOF-physics weights are both $10^{-5}$. The natural frequency is initialized at \(2\pi(2)\ \mathrm{rad/s}\) and optimized during training, and the damping ratio is initialized at 0.05 and constrained to $[0.005,\,0.30]$. Development training allows up to 4000 epochs, with the final refit epoch count determined from the development folds.

The experiment-only PINN receives no simulation response and gives a mean $R^2$ of 0.5027, a mean NRMSE of \SI{7.6414}\%, and a mean NMAE of \SI{4.0873}\%. The TDNN-assisted residual model gives a mean $R^2$ of 0.6515, a mean NRMSE of \SI{6.4180}\%, and a mean NMAE of \SI{3.1637}\%. The proposed discrepancy-correction formulation gives a mean $R^2$ of 0.8197, a mean NRMSE of \SI{4.6055}\%, and a mean NMAE of \SI{1.9297}\%. Table~\ref{tab:performance} summarizes these aggregate values. Both simulation-informed cases have numerically higher mean $R^2$ than the experiment-only PINN. However, differences in preprocessing, architecture, training budget, and selection rules mean that these values describe the staged study outcome rather than a controlled model ranking.

\subsection{TDNN-Assisted Residual Result} \label{sec:correction_results}

The TDNN-assisted residual model combines the TDNN baseline with the observed residual $\mathbf{r}^{\mathrm{e}}_c$ defined in Eq.~\eqref{eq:observation_decomposition}. The model is trained in two stages. For this comparison case, the TDNN stage follows the same simulation-based baseline construction described in  Sections~\ref{sec:tdnn} and~\ref{sec:tdnn_align}. The TDNN uses a 40-sample simulation-response history and is trained to reproduce the simulation displacement over the development conditions. After training the TDNN, its parameters are frozen, and the development-fitted amplitude factor $\rho^{\star}$ is applied to obtain the aligned baseline $\widetilde{\mathbf{q}}^{\mathrm{ML}}_c$.

A separate data-driven residual network is then trained to reproduce the observed residual $\mathbf{r}^{\mathrm{e}}_c$ in Eq.~\eqref{eq:observation_decomposition}. It receives the time $t_i$, speed $v_c$, and aligned TDNN baseline $\widetilde{\mathbf{q}}^{\mathrm{ML}}_c(t_i)$ as inputs. The network contains four hidden layers with 96 neurons per layer and uses hyperbolic-tangent activation functions. Let $\widehat{\mathbf{r}}^{\mathrm{NN}}_c(t_i;\boldsymbol{\eta})$ denote its predicted
four-channel residual. Its training loss is
\begin{equation}
\begin{gathered}
    \boldsymbol{\eta}^{\star}
    \in\operatorname*{arg\,min}_{\boldsymbol{\eta}}\mathcal{L}_{\mathrm{resNN}}(\boldsymbol{\eta}),\\
    \mathcal{L}_{\mathrm{resNN}}(\boldsymbol{\eta})
    :=
    \frac{1}{N_{\mathrm{res}}}
    \sum_{c=1}^{4}
    \sum_{i=i^-_{\mathrm{res},c}}^{i^+_{\mathrm{res},c}}
    \left\|
    [\mathbf{S}_q]^{-1}
    \left(
    \widehat{\mathbf{r}}^{\mathrm{NN}}_c(t_i;\boldsymbol{\eta})
    -
    \mathbf{r}^{\mathrm{e}}_c(t_i)
    \right)
    \right\|_2^2.
\end{gathered}
\label{eq:tdnn_nn_residual_loss}
\end{equation}
The corrected response is obtained by 
\begin{equation}
    \widehat{\mathbf{q}}^\mathrm{resNN}_c(t_i)
    :=\widetilde{\mathbf{q}}^{\mathrm{ML}}_c(t_i)+\widehat{\mathbf{r}}
    _{c}^\mathrm{NN}(t_i;\boldsymbol{\eta^*})\in\mathbb{R}^{J}.
\end{equation}
Unlike the proposed discrepancy-correction formulation, this comparison model uses no acceleration target, reduced dynamic-balance loss, or smoothness loss.

The main training hyperparameters were selected through a separate development-only hyperparameter search. The search varied the common $L_2$ weight-decay coefficient, the TDNN and residual-network learning rates, and the number and width of the residual-network hidden layers. Candidate configurations were evaluated by leave-one-speed-out cross-validation over the 300--360~km/h development conditions and ranked by the mean validation RMSE of the corrected displacement. The 385~km/h condition was not used in this selection. The selected configuration uses learning rates of $3\times10^{-4}$ and $10^{-2}$ for the TDNN and residual network, respectively, a common weight decay of $10^{-6}$, and four residual-network hidden layers with 96 neurons per layer. Both networks are trained with the Adam optimizer. Development training allows up to 600 epochs for the baseline TDNN and 1000 epochs for the residual NN, with the final refit epoch count for each model determined from the development folds.

It gives a numerically higher mean $R^2$ (0.6515 versus 0.5027) and lower mean NRMSE (\SI{6.4180}\% versus \SI{7.6414}\%) and mean NMAE (\SI{3.1637}\% versus \SI{4.0873}\%) than the experiment-only PINN. These numerical differences belong only to that comparison.

\begin{figure}[H]
	\centering
    \begin{subfigure}[t]{0.48\textwidth}
        \centering
        \includegraphics[width=\linewidth]{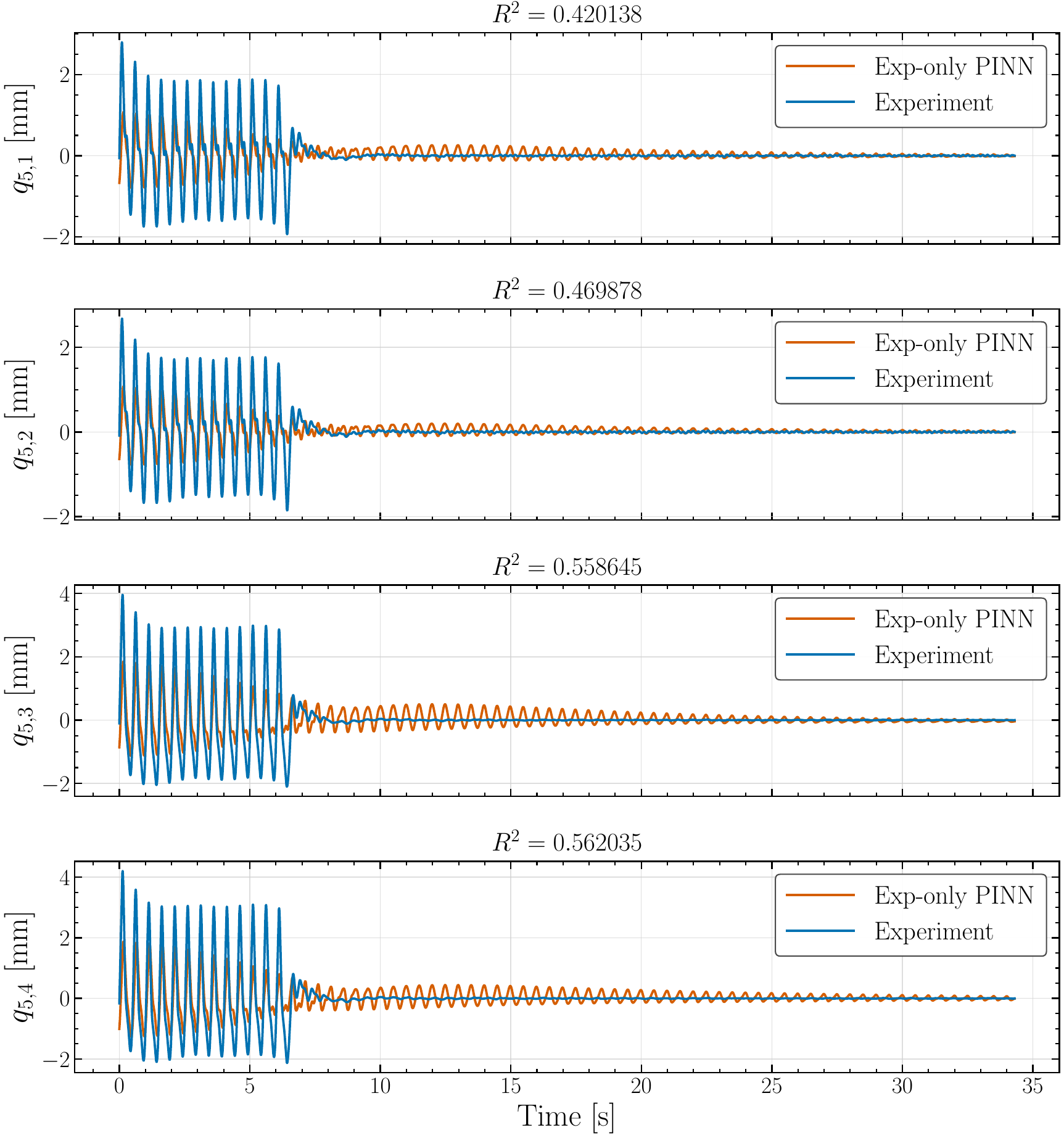}
        \caption{Experiment-only PINN}
        \label{fig:response_case_exp_only}
    \end{subfigure}
    \hfill
    \begin{subfigure}[t]{0.48\textwidth}
        \centering
        \includegraphics[width=\linewidth]{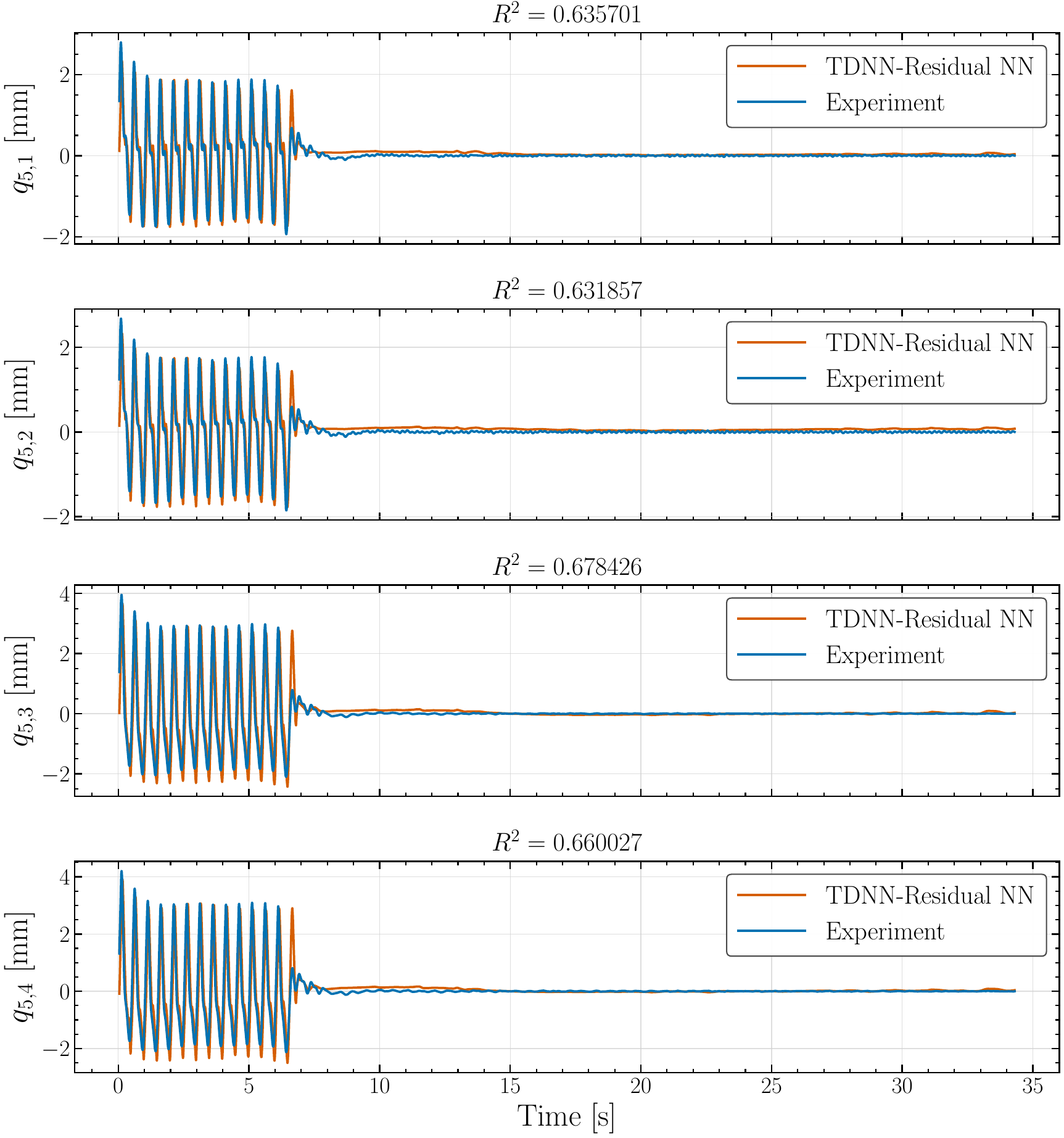}
        \caption{TDNN-assisted residual model}
        \label{fig:response_case_tdnn_residual}
    \end{subfigure}
    \hfill
    \begin{subfigure}[t]{0.48\textwidth}
        \centering
        \includegraphics[width=\linewidth]{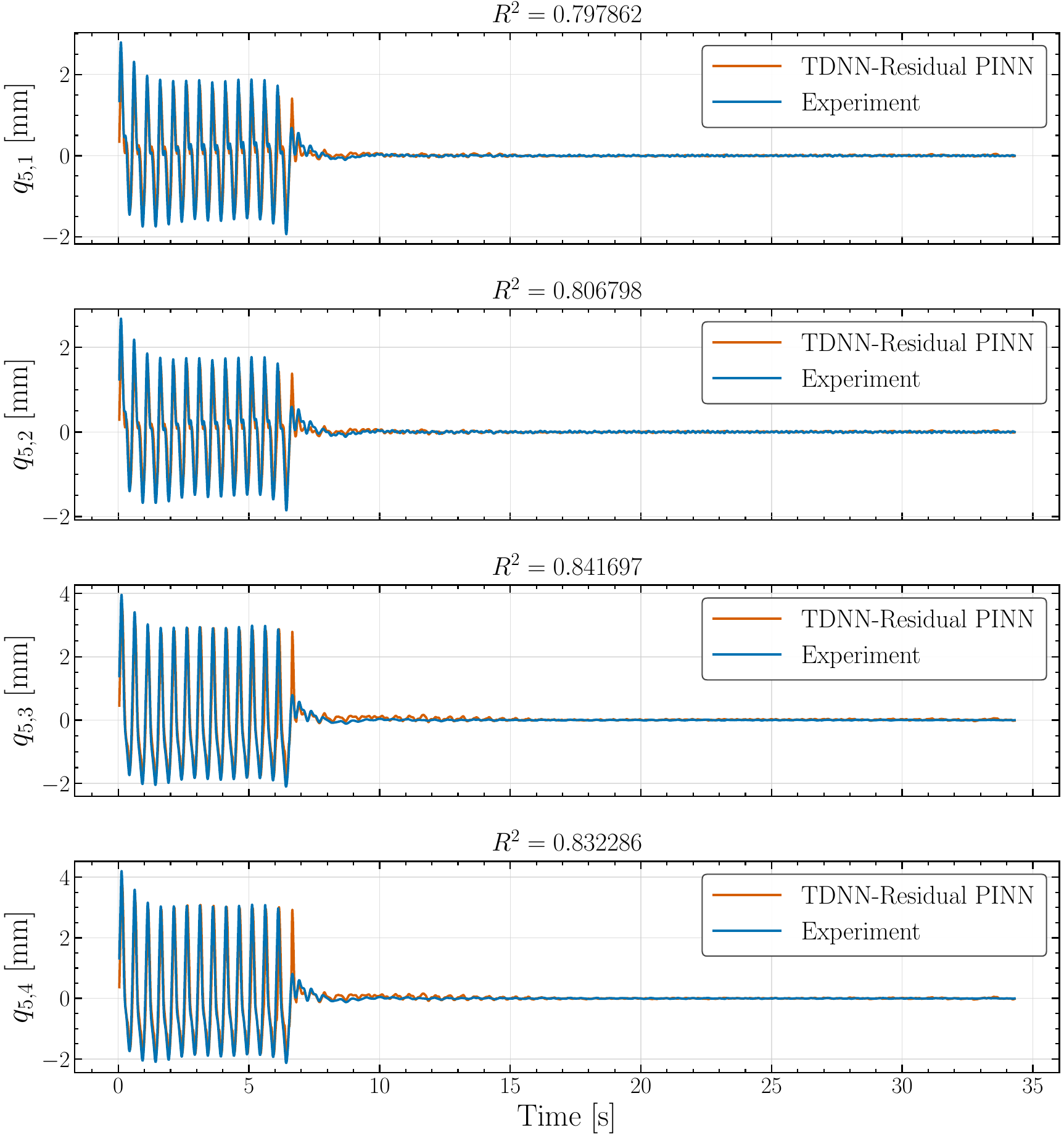}
        \caption{Proposed discrepancy-correction formulation}
        \label{fig:response_case_proposed}
    \end{subfigure}
	\caption{Response comparison at the selected condition: measured response and predictions from (a) the experiment-only PINN, (b) the TDNN-assisted residual model, and (c) the proposed discrepancy-correction formulation. Each panel presents the four FW, FB, RW, and RB channel histories over the same interval with common axes, enabling direct comparison of phase, amplitude, and post-excitation behavior.}
	\label{fig:response_case_overlays}
\end{figure}

\begin{figure}[H]
	\centering
	\includegraphics[width=0.98\linewidth]{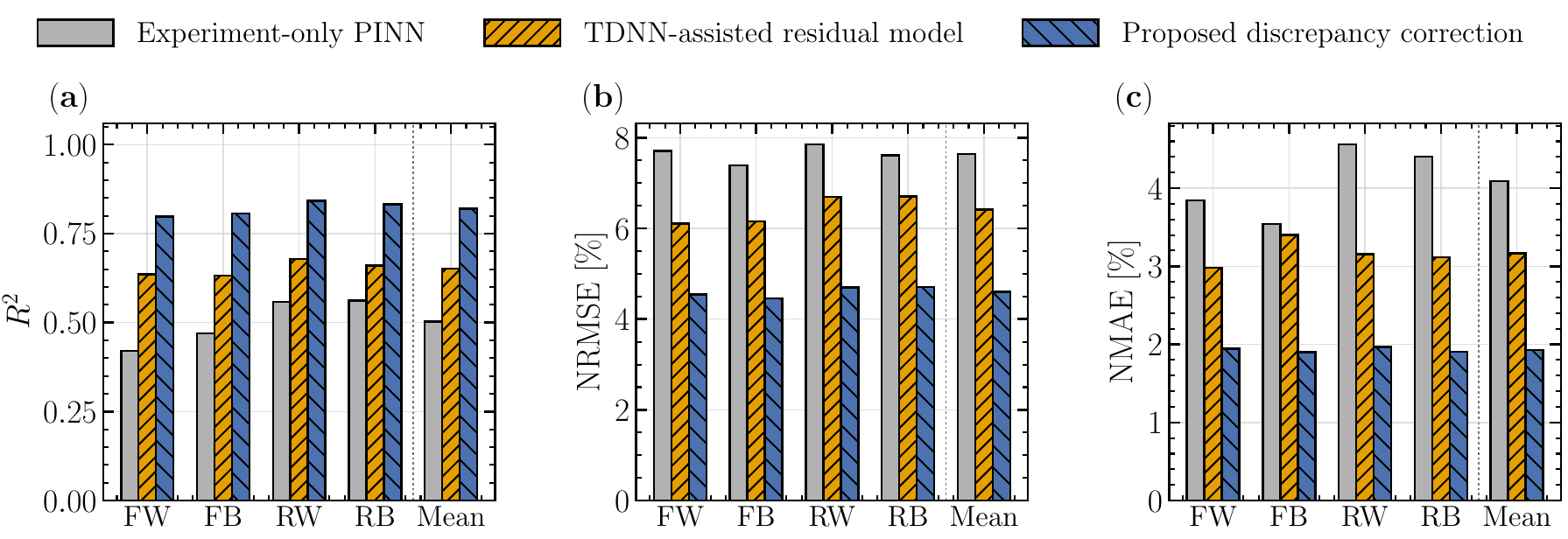}
	\caption{Channelwise accuracy of the experiment-only PINN, TDNN-assisted residual model, and proposed discrepancy-correction formulation: (a) $R^2$, (b) NRMSE, and (c) NMAE for the FW, FB, RW, and RB responses. The channel values are presented individually, and the arithmetic mean reported in Table~\ref{tab:performance} is included separately as a reference.}
	\label{fig:channelwise_metrics}
\end{figure}

Table~\ref{tab:performance} lists the metrics for the experiment-only PINN, TDNN-assisted residual model, and proposed discrepancy-correction formulation. Figures~\ref{fig:response_case_overlays} and \ref{fig:channelwise_metrics} compare their condition-level response histories and channelwise accuracy measures.

\begin{table}[H]
    \centering
    \caption{Reported aggregate metrics for the response-comparison cases. The normalized errors are expressed relative to the measured channelwise response ranges at the 385~km/h evaluation condition.}
    \label{tab:performance}
    \small
    \begin{tabular}{>{\raggedright\arraybackslash}p{0.45\textwidth}ccc}
        \hline
        \noalign{\vskip 0.6ex}
        Case & Mean $R^2$ &
        \shortstack{Mean NRMSE (\%)} &
        \shortstack{Mean NMAE (\%)} \\
        \hline
        Uncorrected simulation response
        & -0.8263 & 14.5355 & 6.2129 \\
        Experiment-only PINN
        & 0.5027 & 7.6414 & 4.0873 \\
        TDNN-assisted residual model
        & 0.6515 & 6.4180 & 3.1637 \\
        Proposed discrepancy-correction formulation
        & 0.8197 & 4.6055 & 1.9297 \\
        \hline
    \end{tabular}
\end{table}

Table~\ref{tab:performance} includes the uncorrected simulation response as a direct numerical baseline for the four evaluation channels. The multibody assessments in Section~\ref{sec:simulation_results} provide complementary consistency evidence for the related full-vehicle model.

This condition provides one data point for defining where the method may be used. Before simulation can reduce repeated development tests, the study must define acceptable accuracy and uncertainty, evaluate independent conditions, and cover the intended ranges of speed, excitation, loading, suspension state, contact geometry, and track or roller boundary. Measurements must remain the reference for updating and evaluating the predictor, and safety-critical or acceptance decisions must follow the applicable assessment rules.

\subsection{Discrepancy-Corrected Response at the Fifth Speed Condition} \label{sec:latest_delta_result}

The discrepancy-correction analysis produces reconstructed responses for the fifth speed condition. Section~\ref{sec:residual} formalizes the effective-dynamics residual and the combined channel-2 acceleration loss. Figure~\ref{fig:response_case_proposed} presents the measured and corrected displacement histories, and Table~\ref{tab:delta_condition5} presents the corresponding channelwise metrics; their aggregate values are also included in Table~\ref{tab:performance}. We use neutral channel names because a unique physical-axis mapping has not been established for all four channels.

The proposed discrepancy corrector was trained using a unit weight for the displacement-residual data term, $10^{-5}$ for the smoothness term and for each of the three constituent penalties of $\mathcal{L}_{\mathrm{acc}}$ (acceleration-head supervision, displacement-derived kinematic acceleration, and head--kinematic consistency), and $10^{-4}$ for the corrective-source physics residual. The trainable $q_{\mathrm{par}}$ coefficients were additionally assigned an $L_2$-regularization coefficient of $10^{-8}$. The correction network $G_\theta$ comprised four hidden layers along the displacement-correction path, with widths of 96, 96, 64, and 64 neurons and hyperbolic-tangent activation functions. The first three layers formed a shared feature extractor, from which a separate linear scalar head predicted the channel-2 acceleration; the displacement-correction branch concatenated this prediction with the shared features and mapped them through the final 64-neuron hidden layer to a four-component linear output. All trainable parameters were optimized using Adam with a learning rate of $1.0\times10^{-2}$ and a weight decay of $1.0\times10^{-6}$. Development training allows up to 1500 epochs for the baseline TDNN and 3000 epochs for the residual PINN, with the final refit epoch count for each model determined from the development folds.

\begin{table}[htbp]
    \centering
    \caption{Normalized displacement-reconstruction metrics for the fifth speed condition. The RMSE and MAE are normalized by the measured response range of each channel.}
    \label{tab:delta_condition5}
    \begin{tabular}{lccc}
        \hline
        Response & $R^2$ & NRMSE (\%) & NMAE (\%) \\
        \hline
        Channel 1 & 0.7979 & 4.5498 & 1.9451 \\
        Channel 2 & 0.8068 & 4.4612 & 1.9009 \\
        Channel 3 & 0.8417 & 4.7009 & 1.9678 \\
        Channel 4 & 0.8323 & 4.7102 & 1.9049 \\
        Mean      & 0.8197 & 4.6055 & 1.9297 \\
        \hline
    \end{tabular}
\end{table}

All four corrected responses achieve $R^2>0.79$ at the fifth speed condition, with a mean $R^2$ of 0.8197. The mean NRMSE and NMAE are \SI{4.6055}\% and \SI{1.9297}\%. Table~\ref{tab:performance} includes these aggregate values to show the staged study outcome. Because alignment, loss terms, and selection procedures were not established across all cases, numerical differences between its rows do not quantify the isolated improvement of the proposed formulation. The held-out fifth condition provides initial evidence of extrapolative prediction beyond the development-speed range. However, one evaluation condition is not sufficient to establish generalization over a broader operating domain.

\subsection{Post-Hoc Oracle Temporal Alignment} \label{sec:post_hoc_oracle}

The response-reconstruction results presented in Section~\ref{sec:latest_delta_result} are obtained without any temporal adjustment based on information from the evaluation condition. In a practical extrapolation setting, the experimental response at the target operating condition is not available. Therefore, the remaining temporal offset cannot be identified by directly comparing the prediction with the measurement. To examine how well the predicted waveform could agree with the measurement under ideal temporal alignment, we additionally report a post-hoc oracle-aligned result for the 385~km/h evaluation condition.

Let $\mathcal{K}$ denote a prespecified finite set of candidate integer sample lags. For each $k\in\mathcal{K}$, let $\mathcal{I}_5(k)$ denote the indices for which both the measured response and the shifted prediction are available, and let $N_5(k):=|\mathcal{I}_5(k)|$. A single global lag is selected for all four response channels as
\begin{equation}
k_{\mathrm{oracle}}^{\star}
\in
\operatorname*{arg\,min}_{k\in\mathcal{K}}
\frac{1}{N_5(k)}
\sum_{i\in\mathcal{I}_5(k)}
\left\|
[\mathbf{S}_{q}]^{-1}
\left(
\mathbf{q}^{\mathrm{e}}_5(t_i)
-
\widehat{\mathbf{q}}_5(t_{i-k})
\right)
\right\|_2^2 .
\label{eq:oracle_time_alignment}
\end{equation}

The corresponding oracle-aligned response is
\begin{equation}
\widehat{\mathbf{q}}^{\mathrm{oracle}}_5(t_i)
:=
\widehat{\mathbf{q}}_5
\left(t_{i-k_{\mathrm{oracle}}^{\star}}\right), i\in\mathcal{I}_5(k^*_\mathrm{oracle}).
\label{eq:oracle_aligned_response}
\end{equation}

\begin{figure}[htbp]
	\centering
	\includegraphics[width=0.85\linewidth]{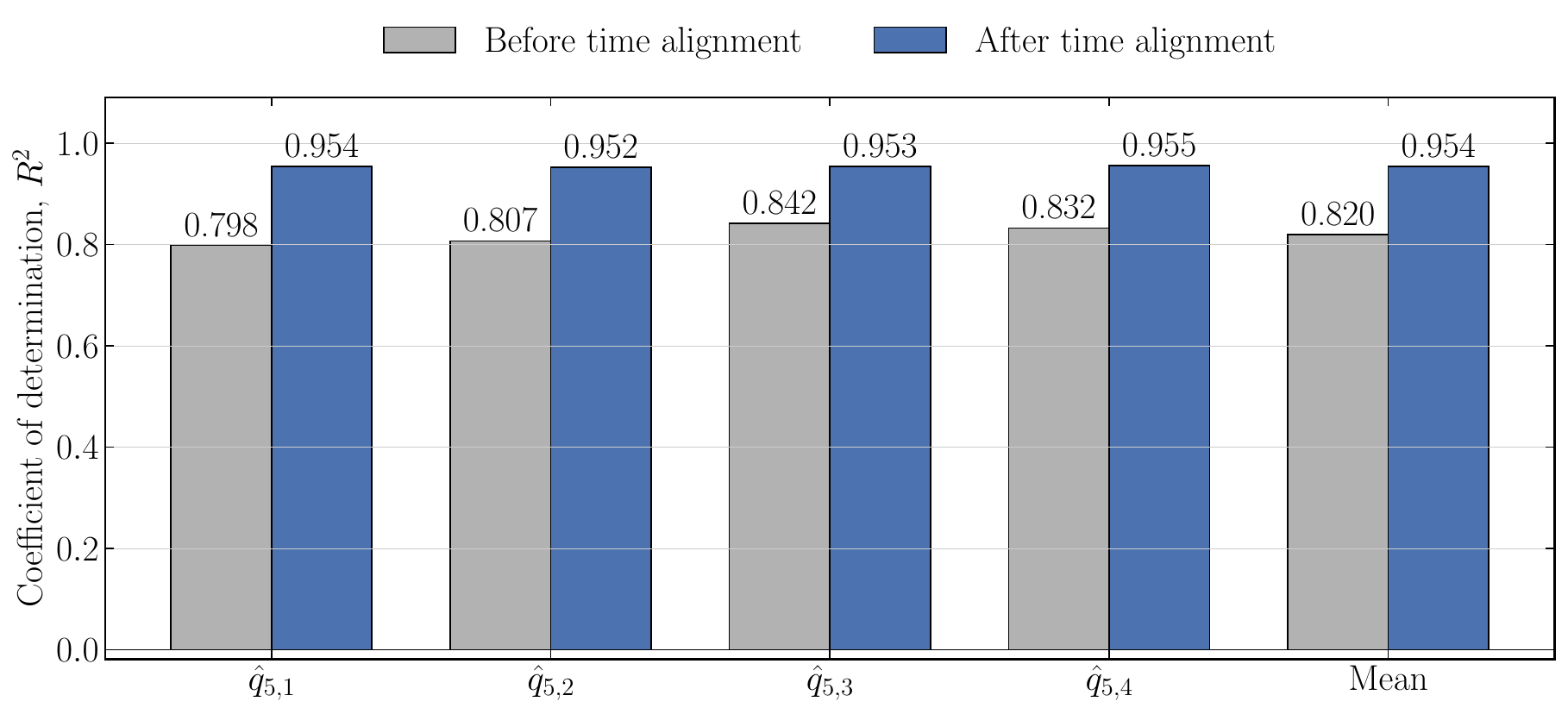}
	\caption{Channelwise and mean $R^2$ values at the 385~km/h evaluation condition before and after post-hoc oracle temporal alignment. The aligned results represent a conditional best-case reference obtained using the measured evaluation response.}
	\label{fig:oracle_time_alignment}
\end{figure}

Figure~\ref{fig:oracle_time_alignment} illustrates the time-lag optimization result for the 385~km/h evaluation condition. For the evaluated condition, the corresponding $R^2$ increases from 0.8197 without target-informed temporal adjustment to 0.9536 after oracle alignment. However, the oracle-aligned value is not treated as a deployable prediction metric because $k_{\mathrm{oracle}}^{\star}$ is determined using the measured response at the evaluation condition.

Instead, this result is used as a conditional best-case reference. It shows the level of agreement that could be obtained if the remaining global temporal offset were known exactly. The large increase in $R^2$ also suggests that a substantial part of the remaining mismatch is related to temporal misregistration, rather than to waveform-shape error alone. For this reason, the oracle-aligned result is reported separately from the primary model-performance comparison.

\subsection{Displacement--Acceleration Channel Diagnostic} \label{sec:acceleration}

We compare the second derivative of measured displacement with the nominally paired measured acceleration at five speeds. Channel 2 shows the strongest repeated association. At the intended development speeds, its direct correlation ranges from 0.2632 to 0.8088 and increases to 0.6609--0.8134 after the lag search. The other three channels remain weak after lag adjustment. At \SI[per-mode=symbol]{300}{\kilo\metre\per\hour}, the normalized spectral-overlap value is 0.2640 for channel 1, 0.8230 for channel 2, 0.3270 for channel 3, and 0.1980 for channel 4. We therefore restrict the proposed acceleration loss to channel 2 and do not formulate it for the other pairs. This selection indicates consistency under the analyzed processing; it does not identify a sensor fault, prove exact kinematic equivalence, or establish the isolated contribution of the combined acceleration loss to the corrected response.

\ref{app:channel_diagnostic} presents the scatterplots, lag-adjusted histories, normalized spectra, and complete numerical values. The fifth speed appears in the diagnostic tables, but channel selection must rely on the development speeds alone if that condition is to serve as an independent evaluation.

\section{Conclusions and Outlook} \label{sec:conclusion}

We developed a multifidelity correction method for railway-bogie response prediction. Its principal methodological contribution is the combined use of low-fidelity MBD histories, high-fidelity physical measurements, and an effective dynamic-balance constraint to learn the response component that the simulation does not explain. This integration extends railway multifidelity modeling beyond the fusion of numerical models of different complexity and gives the experiment-informed correction an explicit model-discrepancy interpretation. The method uses a TDNN and development-fitted alignment to construct a condition-dependent baseline, distinguishes latent model discrepancy from measurement error, and separates simulation-model consistency from response-reconstruction evidence. The correction objective combines residual matching and temporal smoothness with the effective dynamic residual and a combined channel-2 acceleration loss selected using displacement--acceleration consistency evidence. The multibody assessments establish limited assembly, numerical, and cross-code consistency. The TDNN-assisted residual model and proposed discrepancy-correction formulation give higher aggregate agreement than the experiment-only PINN, although the staged comparison does not isolate the contribution of each method component. For the evaluated reconstruction case, the corrected response reaches a mean $R^2$ of 0.8197, a mean NRMSE of \SI{4.6055}\%, and a mean NMAE of \SI{1.9297}\%. A response-consistency diagnostic supports the selective use of the combined acceleration loss, but controlled ablation remains necessary to quantify its isolated contribution.

The effective $[\Delta\mathbf{M}]$, $[\Delta\mathbf{C}]$, and $[\Delta\mathbf{K}]$ terms describe discrepancy in a reduced dynamic model; they do not identify physical component errors from four sensor histories. The held-out 385~km/h condition provides initial evidence of extrapolative response prediction beyond the development-speed range, with the uncorrected simulation response included as a channel-matched numerical baseline. However, a single evaluation condition is not sufficient to establish generalization across broader speed, excitation, loading, suspension, or boundary-condition ranges. Future work should therefore extend the independent evaluation set, perform controlled loss ablations, verify the sensor-to-coordinate map, and quantify prediction uncertainty over the intended operating domain. Physical tests remain the reference for safety, certification, and vehicle acceptance.

\section*{Acknowledgements}

This research was supported by a grant from R\&D Program (PK26111A0) of the Korea Railroad Research Institute and by the Korea Agency for Infrastructure Technology Advancement (KAIA) grant funded by the Ministry of Land, Infrastructure and Transport (Grant RS-2024-00417481)

\appendix
\setcounter{table}{0}
\setcounter{figure}{0}
\section{Selected Physics-Based Model Parameters} \label{app:model_parameters}

This appendix provides selected numerical parameters and implementation details of the full-vehicle model. The rigid-body properties used in the model are reported in Table~\ref{tab:appendix_rigid_bodies}. Detailed suspension and connection parameters, attachment coordinates, and nonlinear damper force--velocity characteristics are not reported due to confidentiality and security restrictions associated with the vehicle model. The component topology and corresponding multibody representations are described in Section~\ref{sec:simulation_model} and Table~\ref{tab:component_implementation}.


\begin{table}[H]
	\centering
	\caption{Rigid-body properties of the full-vehicle model.}
	\label{tab:appendix_rigid_bodies}
	\small
	\begin{tabular}{lcccc}
		\hline
		Component & Count & 
		\shortstack{Center of mass\\(m)} & 
		\shortstack{Mass\\(Mg)} &
		\shortstack{Mass moments\\(Mg\,m$^2$)} \\
		\hline
		Car body & 1 & $(0.200,-0.013,1.642)$ & 34.299 & $(85.3,1598,1598)$ \\
		Bogie frame & 2 & $(-0.014,0,0.566)$ & 4.343 & $(2.33,2.99,5.05)$ \\
		Wheelset & 4 & $(0,0,0.430)$ & 1.936 & $(1.01,0.01,1.01)$ \\
		\hline
	\end{tabular}
\end{table}

The center-of-mass triplet lists the longitudinal, lateral, and vertical components. The mass-moment triplet lists moments about the longitudinal, lateral, and vertical axes. The car-body coordinates are measured from the car-body reference, while the bogie-frame and wheelset coordinates are measured from the local bogie reference.

The rod-bush model contains a series branch and a parallel branch separated by an internal node. Along the longitudinal or vertical branch, its externally transmitted force is
\begin{equation}
	F_{\mathrm{ext}}
	=
	\frac{k_{\mathrm{par}}k_{\mathrm{ser}}}{k_{\mathrm{par}}+k_{\mathrm{ser}}}d_b
	+
	\frac{k_{\mathrm{ser}}}{k_{\mathrm{par}}+k_{\mathrm{ser}}}F_{0},
	\label{eq:rod_bush_preload}
\end{equation}
where $F_{\mathrm{ext}}\in\mathbb{R}$ is the external marker force, $F_{0}\in\mathbb{R}$ is the internal preload, $k_{\mathrm{ser}}, k_{\mathrm{par}}\in\mathbb{R}^{+}$ are the series and parallel stiffnesses, and $d_b\in\mathbb{R}$ is the total relative deformation along the element direction. When $d_b=0$, Eq.~\eqref{eq:rod_bush_preload} shows that the externally transmitted force is generally a stiffness-weighted fraction of the internal preload rather than the preload itself. The preload comparison therefore uses the external marker force rather than the internal preload value.

\setcounter{table}{0}
\setcounter{figure}{0}
\section{Displacement--Acceleration Channel-Consistency Evidence} \label{app:channel_diagnostic}

This appendix assesses whether each measured acceleration is consistent with the second derivative of its nominally paired measured displacement under the selected processing. The comparisons do not validate either sensor independently.

\subsection{Direct time-domain comparison}

\begin{figure}[H]
	\centering
	\begin{subfigure}[t]{0.47\linewidth}
		\includegraphics[width=\linewidth]{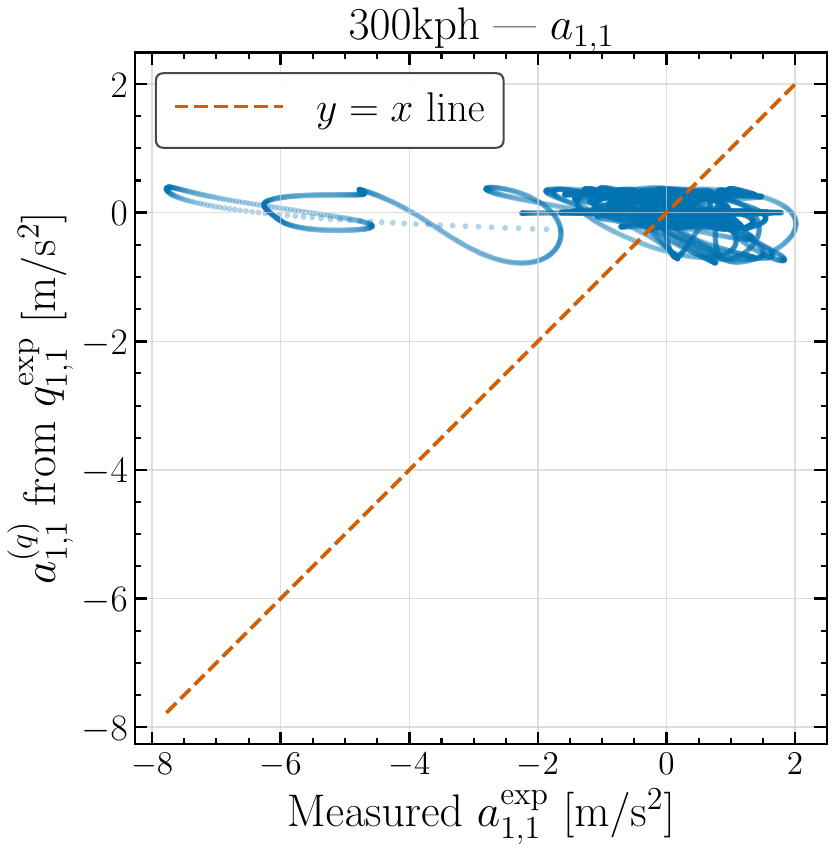}
		\caption{Channel 1.}
	\end{subfigure}\hfill
	\begin{subfigure}[t]{0.47\linewidth}
		\includegraphics[width=\linewidth]{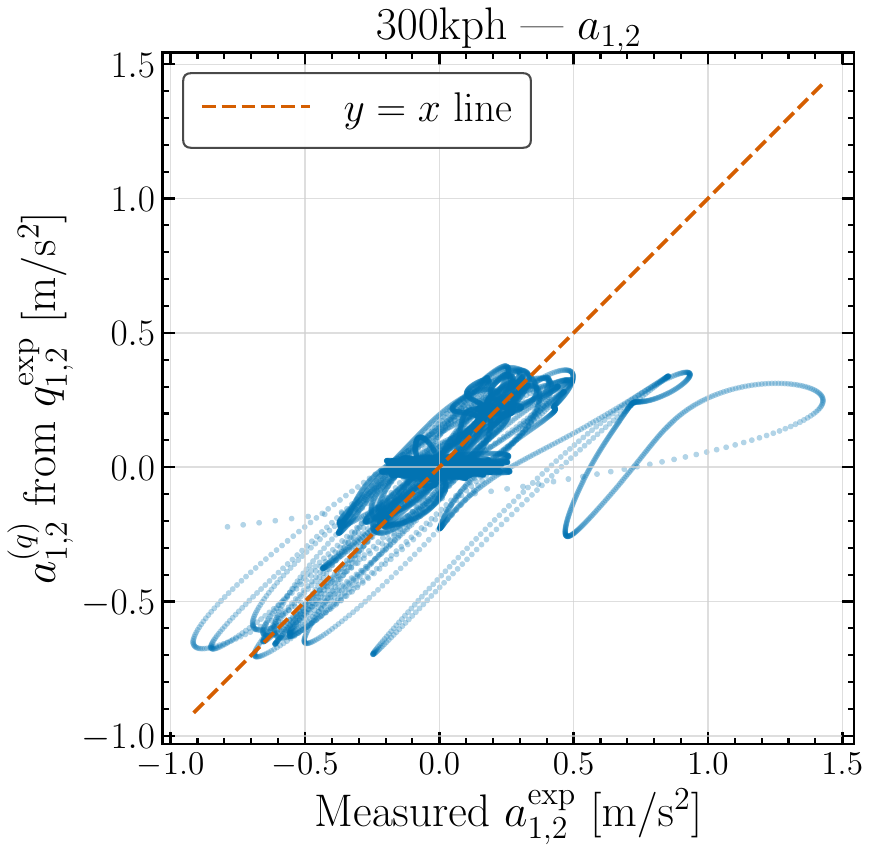}
		\caption{Channel 2.}
	\end{subfigure}
	\begin{subfigure}[t]{0.47\linewidth}
		\includegraphics[width=\linewidth]{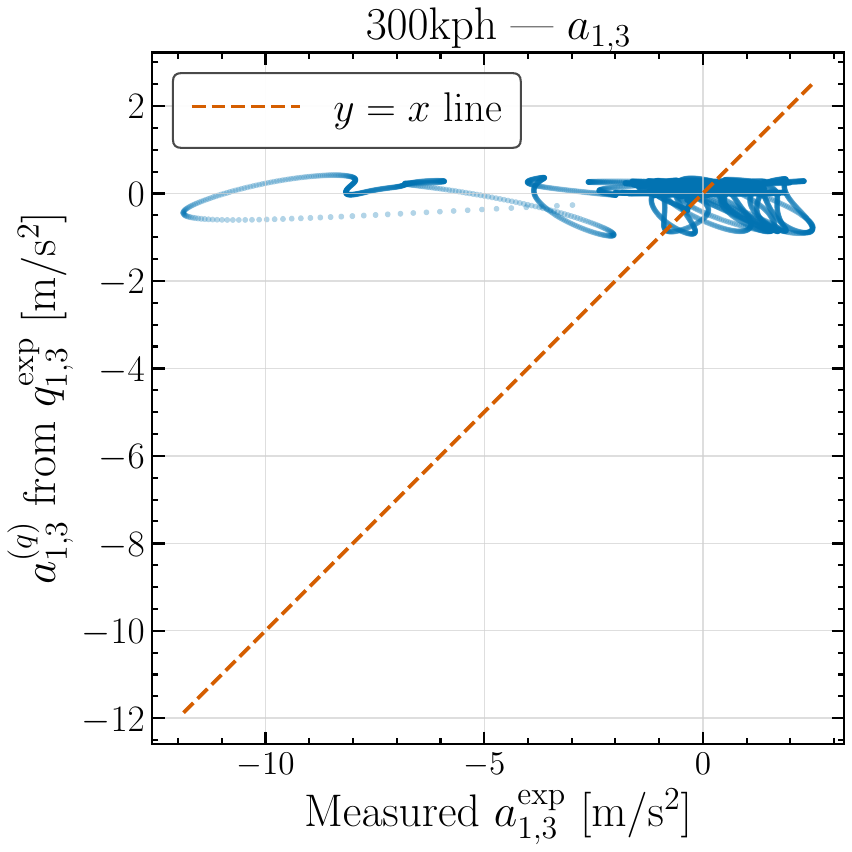}
		\caption{Channel 3.}
	\end{subfigure}\hfill
	\begin{subfigure}[t]{0.47\linewidth}
		\includegraphics[width=\linewidth]{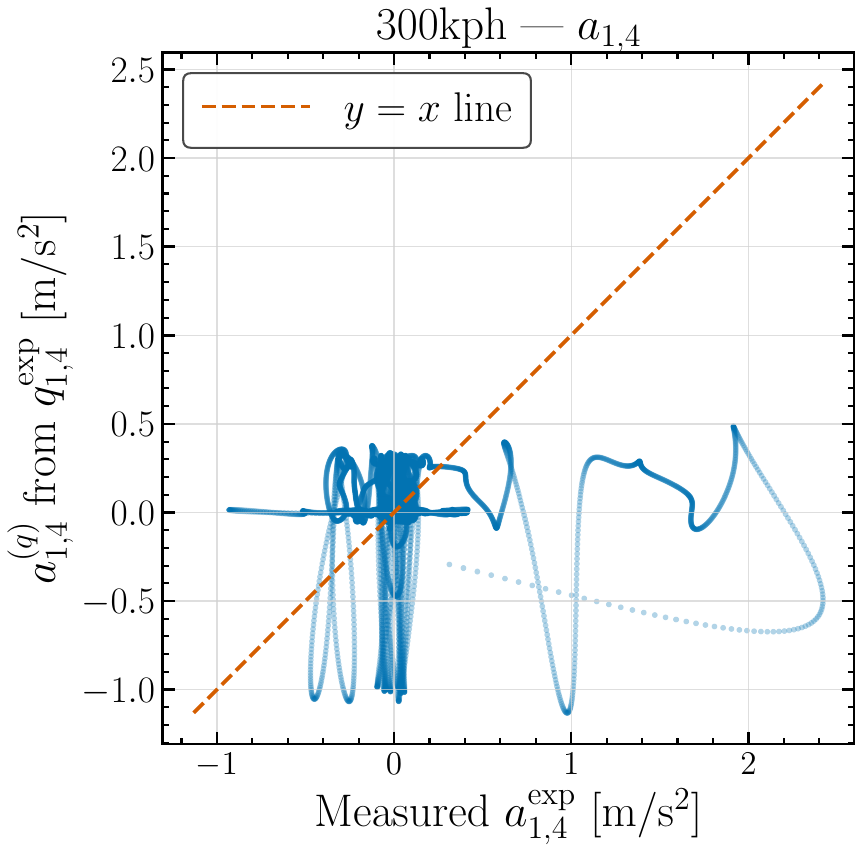}
		\caption{Channel 4.}
	\end{subfigure}
	\caption{Channelwise comparison between the second time derivative of measured displacement and sign-corrected measured acceleration at \SI[per-mode=symbol]{300}{\kilo\metre\per\hour}. The dashed line denotes one-to-one agreement.}
	\label{fig:appendix_scatter}
\end{figure}

\begin{table}[H]
	\centering
	\caption{Direct agreement between displacement-derived and measured acceleration.}
	\label{tab:appendix_direct_acceleration}
	\scriptsize
	\begin{tabular}{crrrrrrrr}
		\hline
		& \multicolumn{2}{c}{Channel 1} & \multicolumn{2}{c}{Channel 2} & \multicolumn{2}{c}{Channel 3} & \multicolumn{2}{c}{Channel 4} \\
		Speed & $R^2$ & Corr. & $R^2$ & Corr. & $R^2$ & Corr. & $R^2$ & Corr. \\
		\hline
		300 & -0.0954 & -0.2204 & 0.5414 & 0.7387 & -0.0616 & -0.1417 & -0.4802 & 0.0206 \\
		320 & -0.0421 & -0.1251 & 0.0593 & 0.2632 & -0.0376 & -0.0608 & -0.2047 & -0.0303 \\
		340 & -0.0335 & -0.0971 & 0.6540 & 0.8088 & -0.0211 & -0.0351 & -1.2691 & 0.0793 \\
		360 & -0.0242 & -0.0502 & 0.6523 & 0.8079 & -0.0126 & -0.0430 & -2.4134 & 0.0196 \\
		385 & -0.0421 & -0.1176 & 0.5481 & 0.7408 & -0.0973 & -0.2191 & -2.7555 & 0.0629 \\
		\hline
	\end{tabular}
\end{table}
The coefficient of determination and Pearson correlation quantify different aspects of agreement and are therefore listed separately. Under Eq.~\eqref{eq:metrics}, a negative $R^2$ indicates that the prediction is less accurate than the sample-mean baseline.

\subsection{Sensitivity to time lag}

\begin{figure}[H]
	\centering
	\begin{subfigure}[t]{0.48\linewidth}
		\includegraphics[width=\linewidth]{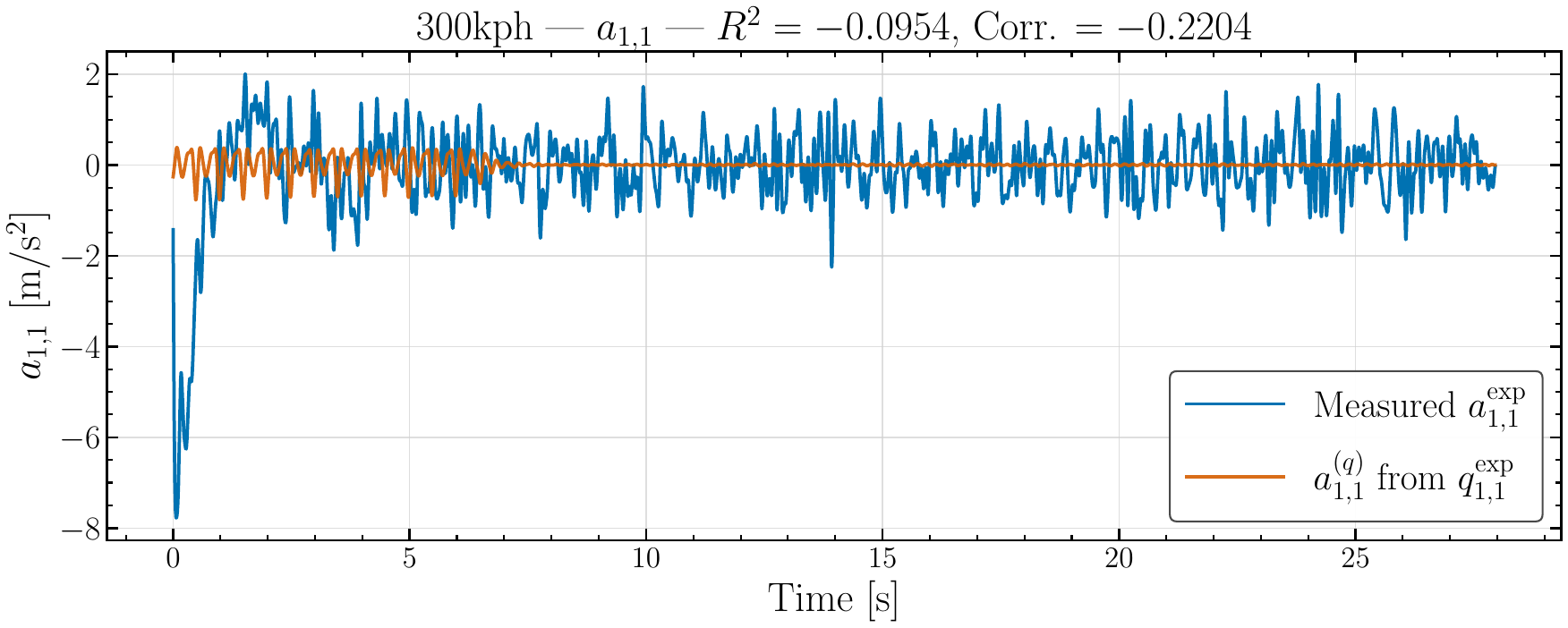}
		\caption{Channel 1.}
	\end{subfigure}\hfill
	\begin{subfigure}[t]{0.48\linewidth}
		\includegraphics[width=\linewidth]{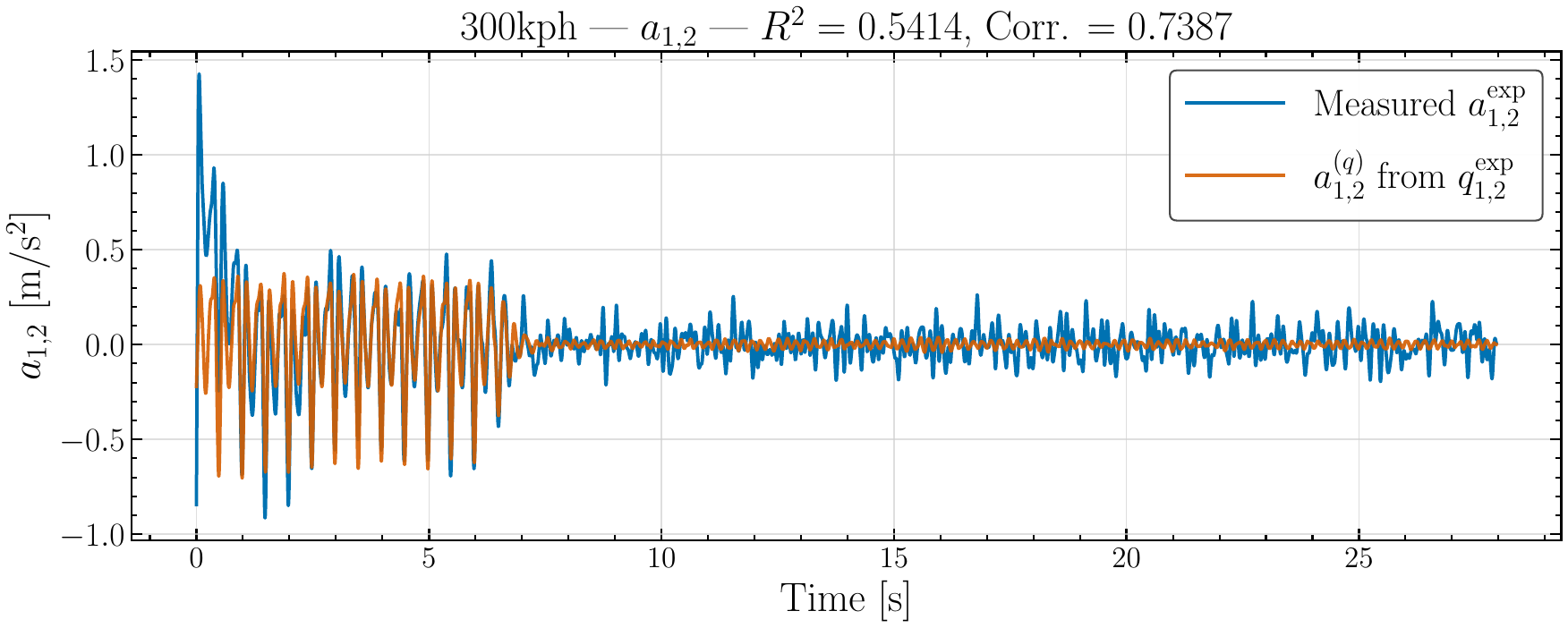}
		\caption{Channel 2.}
	\end{subfigure}
	\begin{subfigure}[t]{0.48\linewidth}
		\includegraphics[width=\linewidth]{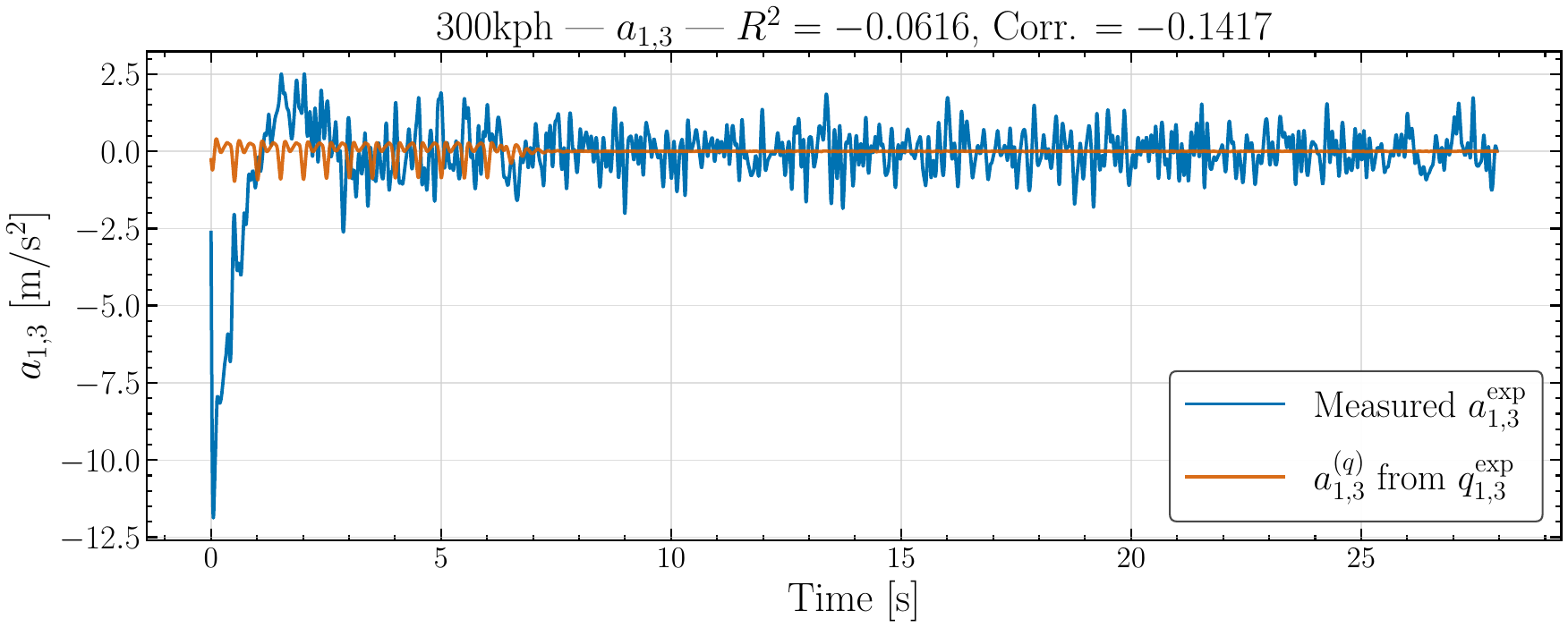}
		\caption{Channel 3.}
	\end{subfigure}\hfill
	\begin{subfigure}[t]{0.48\linewidth}
		\includegraphics[width=\linewidth]{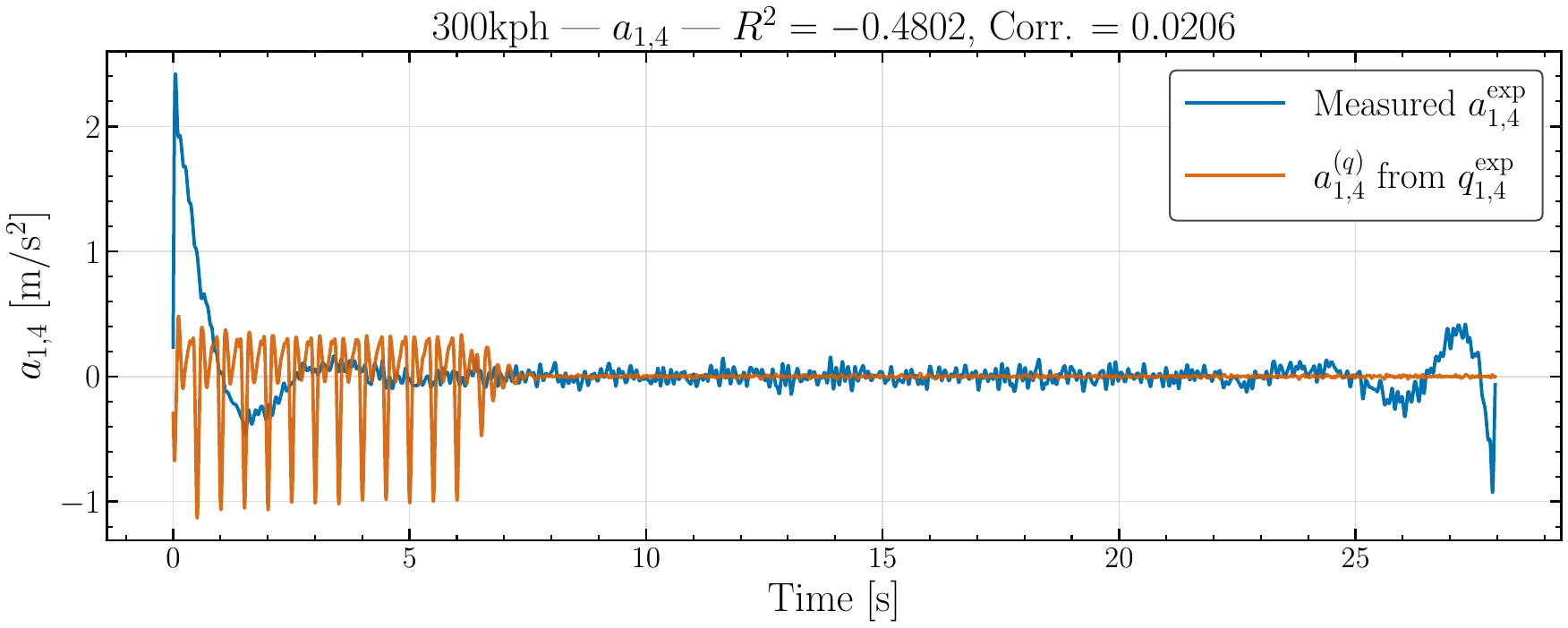}
		\caption{Channel 4.}
	\end{subfigure}
	\caption{Measured acceleration and the second derivative of measured displacement at \SI[per-mode=symbol]{300}{\kilo\metre\per\hour} after the channelwise lag search. The panel titles show the signed lag and correlation.}
	\label{fig:appendix_lag}
\end{figure}

\begin{table}[H]
	\centering
	\caption{Maximum absolute correlation after shifting each sequence by at most 2000 samples.}
	\label{tab:appendix_lag}
	\begin{tabular}{ccccc}
		\hline
		Speed (km/h) & Channel 1 & Channel 2 & Channel 3 & Channel 4 \\
		\hline
		300 & 0.2762 & 0.7978 & 0.2424 & 0.0791 \\
		320 & 0.1563 & 0.6609 & 0.1289 & 0.0816 \\
		340 & 0.1253 & 0.8109 & 0.1577 & 0.1165 \\
		360 & 0.0615 & 0.8134 & 0.1268 & 0.0958 \\
		385 & 0.1179 & 0.7478 & 0.2352 & 0.1593 \\
		\hline
	\end{tabular}
\end{table}
The table contains positive values, whereas the panel titles show negative signed correlations for channels 1, 3, and 4 at \SI[per-mode=symbol]{300}{\kilo\metre\per\hour}. We therefore interpret the tabulated quantities as magnitudes. The sample interval is required before the 2000-sample bound can be stated in seconds.

\subsection{Frequency-domain comparison}

\begin{figure}[H]
	\centering
	\begin{subfigure}[t]{0.48\linewidth}
		\includegraphics[width=\linewidth]{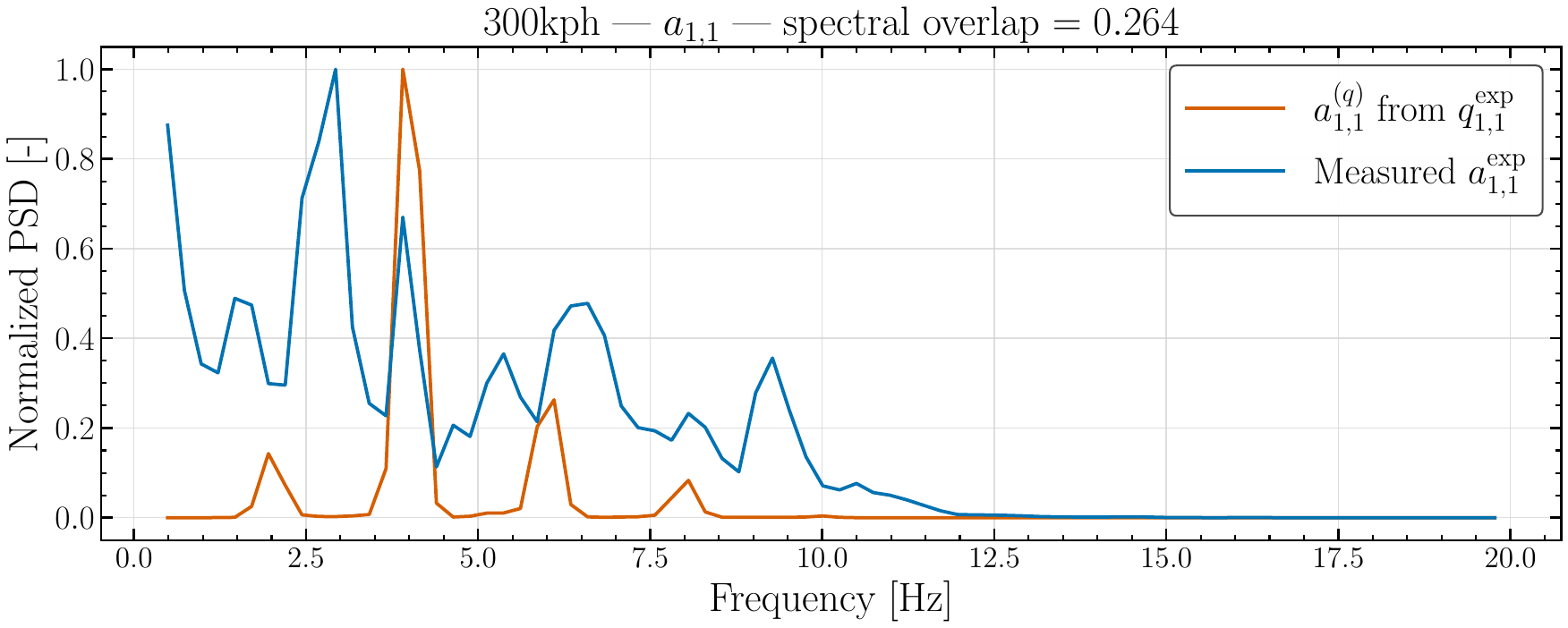}
		\caption{Channel 1.}
	\end{subfigure}\hfill
	\begin{subfigure}[t]{0.48\linewidth}
		\includegraphics[width=\linewidth]{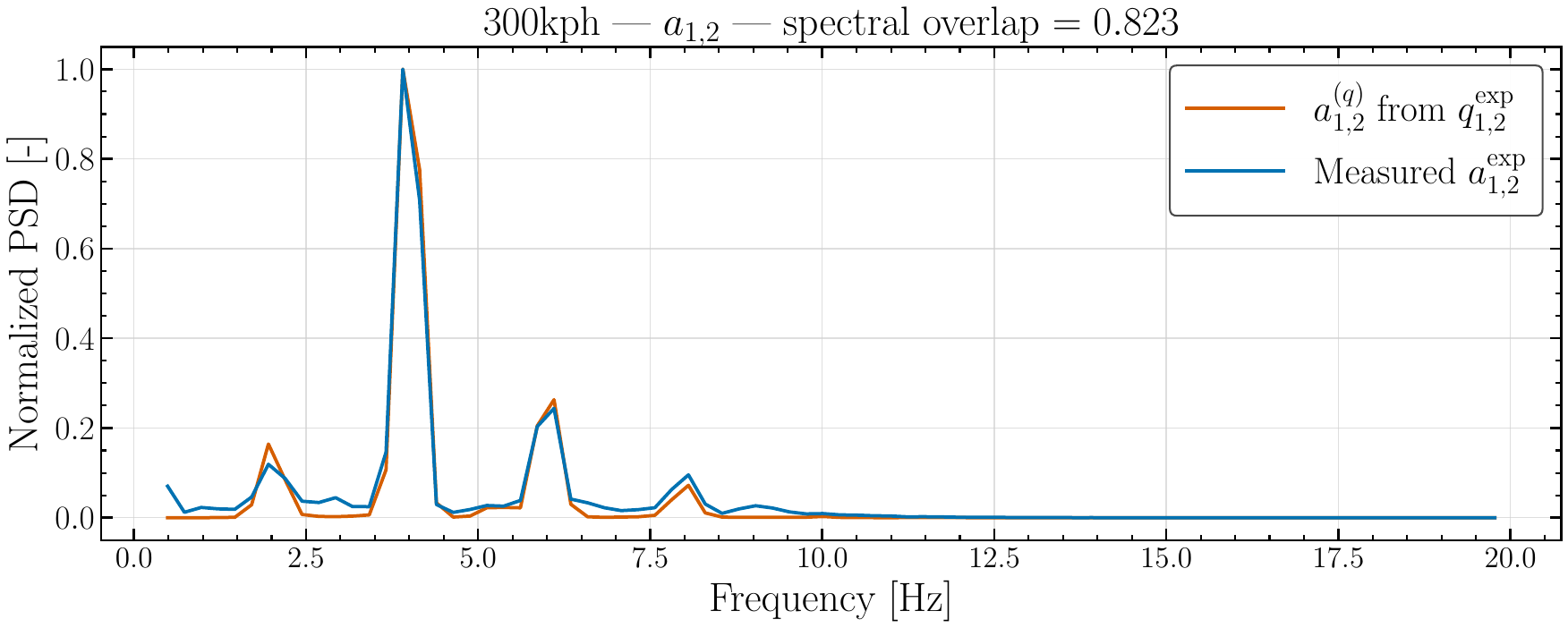}
		\caption{Channel 2.}
	\end{subfigure}
	\begin{subfigure}[t]{0.48\linewidth}
		\includegraphics[width=\linewidth]{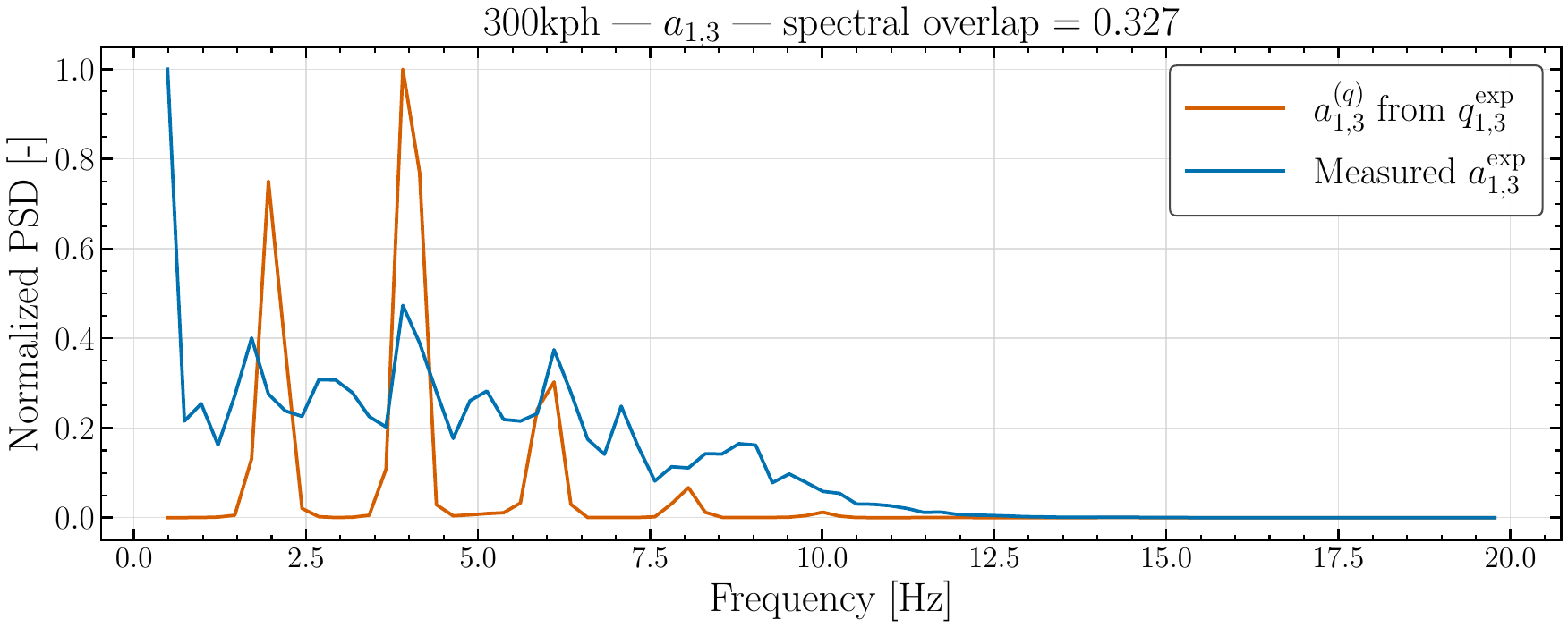}
		\caption{Channel 3.}
	\end{subfigure}\hfill
	\begin{subfigure}[t]{0.48\linewidth}
		\includegraphics[width=\linewidth]{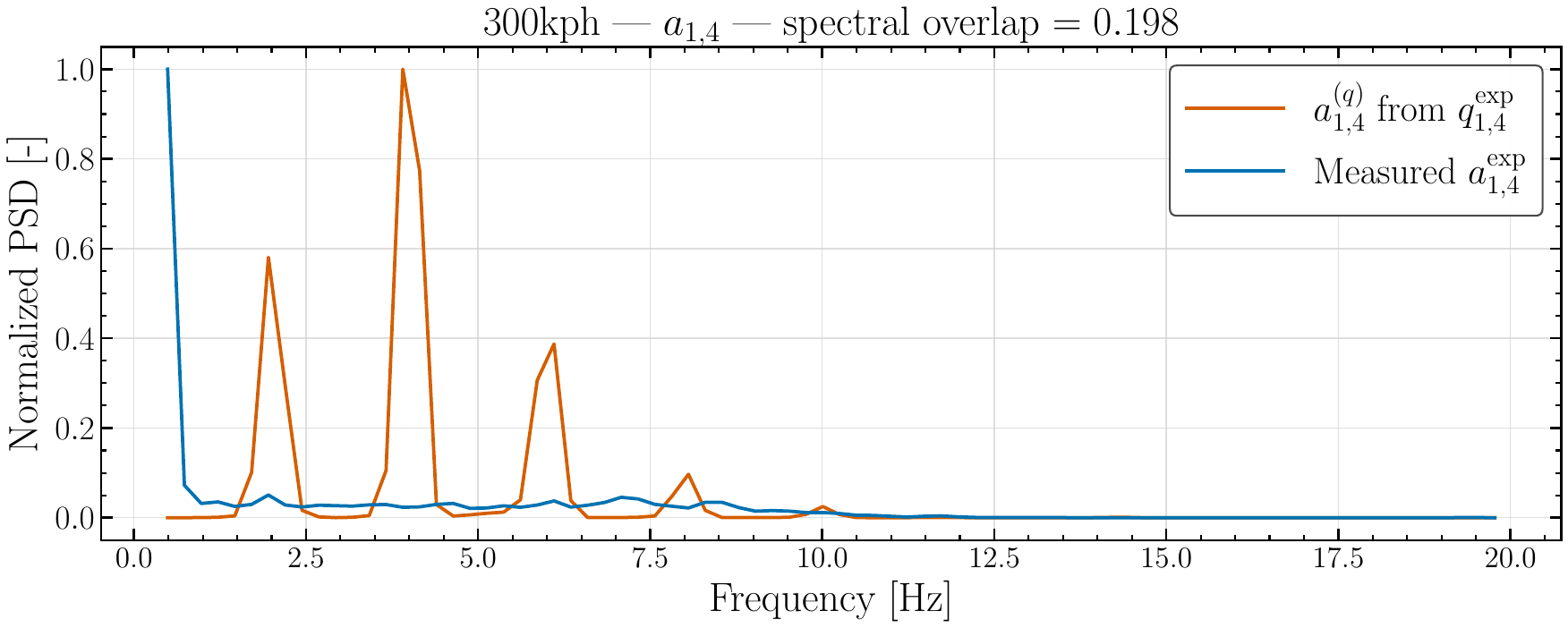}
		\caption{Channel 4.}
	\end{subfigure}
	\caption{Normalized power-spectral-density comparison at \SI[per-mode=symbol]{300}{\kilo\metre\per\hour}. The spectral-overlap value is 0.2640 for channel 1, 0.8230 for channel 2, 0.3270 for channel 3, and 0.1980 for channel 4. These plots show spectral overlap, not magnitude-squared coherence.}
	\label{fig:appendix_spectra}
\end{figure}

\begingroup
\small
\setlength{\emergencystretch}{2em}
\Urlmuskip=0mu plus 1mu\relax
\bibliography{references}
\bibliographystyle{elsarticle-num}
\endgroup

%
\end{document}